\documentclass{article} 
\usepackage{iclr2027_conference,times}

\usepackage{amsmath,amsfonts,bm}

\def\eqref#1{equation~\ref{#1}}

\def\1{\bm{1}}

\DeclareMathAlphabet{\mathsfit}{\encodingdefault}{\sfdefault}{m}{sl}
\SetMathAlphabet{\mathsfit}{bold}{\encodingdefault}{\sfdefault}{bx}{n}

\usepackage{hyperref}
\usepackage{url}
\usepackage{graphicx}
\usepackage{booktabs}
\usepackage{multirow}
\usepackage{makecell}    
\usepackage{subcaption}
\usepackage{wrapfig} 
\usepackage{algorithm}
\usepackage{algpseudocode}
\usepackage{tabularx}
\usepackage{array}

\usepackage[table]{xcolor}

\title{Steering Video Diffusion Transformers with Massive Activations}

\author{Xianhang Cheng$^1$, Yujian Zheng$^{1\dagger}$, Zhenyu Xie$^1$,
Tingting Liao$^1$ \& Hao Li$^{1,2}$ \\
$^1$MBZUAI \quad $^2$Pinscreen \\
}

\iclrfinalcopy
\makeatletter
\def\@oddhead{\vbox{\hbox to \textwidth{\normalfont{Preprint}\hfil}
  \vskip 2pt \hrule height 0.4pt}}
\def\@evenhead{\@oddhead}
\makeatother

\begin{document}

\maketitle
\vspace{-1.5em}
\renewcommand{\thefootnote}{}
\footnotetext{$^\dagger$Corresponding Author. \hfill
\textbf{Project page:} \url{https://xianhang.github.io/webpage-STAS/}}
\renewcommand{\thefootnote}{\arabic{footnote}}

\begin{abstract}
In this work, we study the role of Massive Activations (MAs), which are rare, high-magnitude spikes confined to a few fixed hidden dimensions in video diffusion transformers (DiTs). We uncover a structured positional hierarchy: MA magnitudes peak at first-frame tokens and recur at the spatial boundary tokens of latent frames, with this pattern being most pronounced during early denoising. We trace this organization to an encoding asymmetry of the video VAEs, whose causal temporal padding and zero spatial padding cause the first latent frame and frame borders to carry reduced content load. Elevated MAs consistently align with these lower-content structural positions. To understand their function, we analyze intermediate representations and find that MAs act as implicit rescalers of residual computation: enlarging MAs suppresses the corresponding self-attention and feed-forward updates, while erasing them amplifies these updates. Together, these observations suggest that MAs serves as a token-level rescaler of residual computation, which video DiTs deploy unevenly, placing the strongest damping at the encoding-asymmetric structural positions. Motivated by this native rescaling behavior, we propose \textbf{St}ructured \textbf{A}ctivation \textbf{S}teering (\textsc{\textbf{STAS}}), a training-free technique that steers MAs at the observed structural positions toward a scaled, model-derived reference during early denoising. \textsc{STAS} requires no additional forward passes and consistently improves video quality and temporal coherence across text-to-video models with negligible overhead.
\vspace{-4pt}
\end{abstract}

\section{Introduction}
\label{sec:intro}

Recent advances in video generation have been driven by diffusion- and flow-based frameworks, evolving from U-Net based text-to-video models \cite{makeavideo, imagen, animatediff, svd, videocrafter1, videocrafter2, lumiere, magicvideo} to diffusion transformer (DiT) architectures \cite{cogvideox, hacohen2024ltx, opensora, opensoraplan, jin2025pyramidal, wan, hunyuanvideo, longcat, ma2025step} that scale better with model capacity and training data. Beyond architecture, video quality and temporal coherence are further improved through training-time objectives \cite{videojam, Videolavit, zhang2025frame}, post-training preference optimization \cite{liu2025improving, cheng2025vpo, yang2025ipo, hpsd}, and inference-time guidance \cite{kim2025model, flowmo, freeinit, freqprior, bytheway, unictrl, s2, hyung2025spatiotemporal}, though at the cost of extra training or inference computation.

In parallel, a separate line of research has uncovered striking activation phenomena inside transformers. In large language models, Massive Activations (MAs) are rare hidden-state activations with magnitudes orders of magnitude larger than typical values \cite{sun2024massive}. They are often viewed as the counterpart of attention sinks, tokens that attract disproportionate attention in attention maps \cite{streamingllm}. The few dimensions hosting MAs remain persistently large across most tokens, matching what recent work terms residual sinks \cite{qiu2026unified}. Related outliers have also been studied in Vision Transformers and image DiTs \cite{darcet2024vision, massive, turri2026few}, yet mostly at the dimension level; how MA magnitudes are distributed across tokens, and what role they play in generation, have received little attention. For video DiTs, this raises two questions: \textit{(1) how are MA magnitudes organized over the video-token layout, and (2) what function do they serve during generation?}

\begin{figure}[t]
\centering

\begin{subfigure}[t]{0.24\linewidth}
    \centering
    \includegraphics[width=\linewidth]{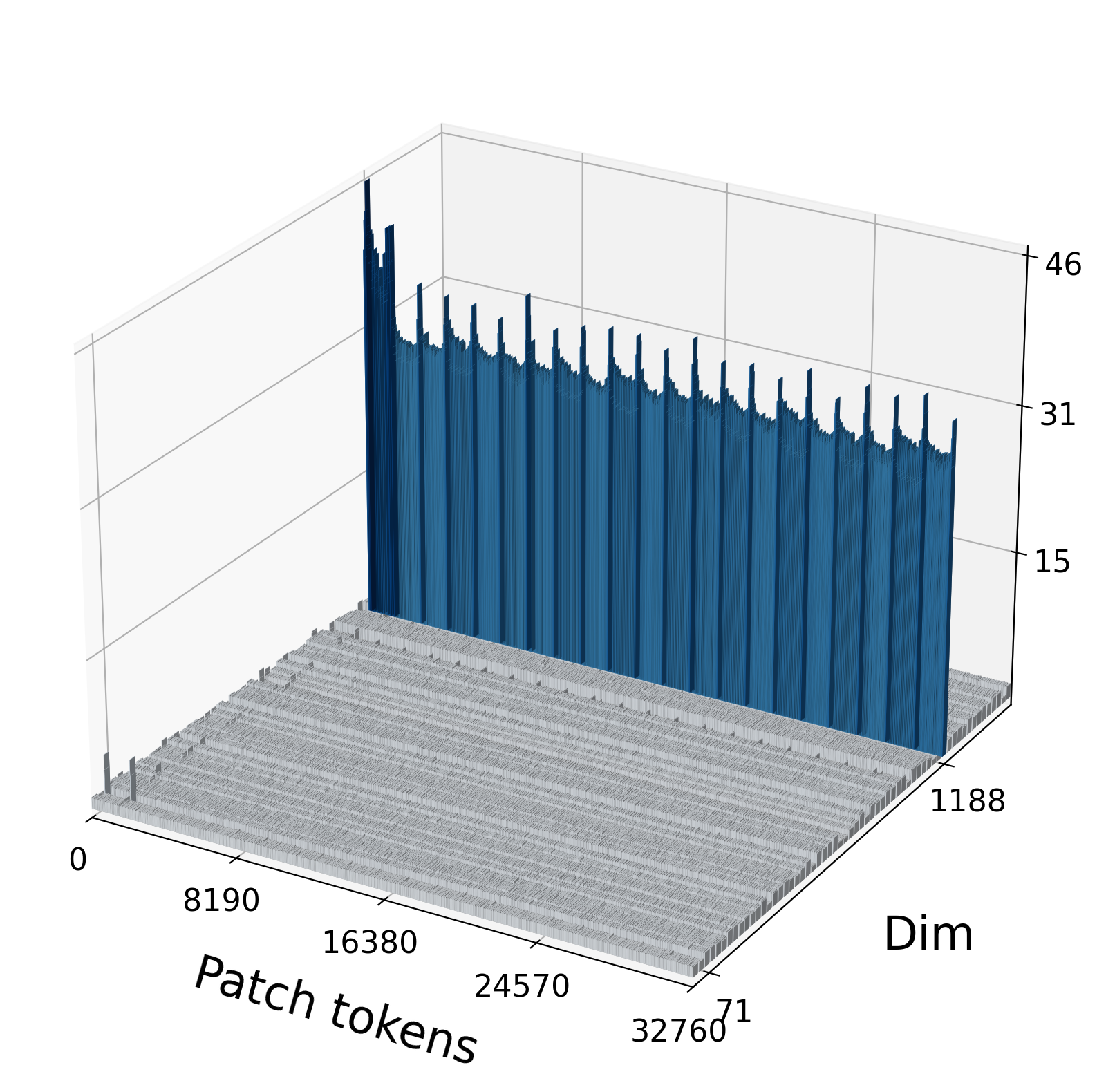}
    \caption{Wan2.1-1.3B}
    \label{fig:fig1a}
\end{subfigure}\hfill
\begin{subfigure}[t]{0.24\linewidth}
    \centering
    \includegraphics[width=\linewidth]{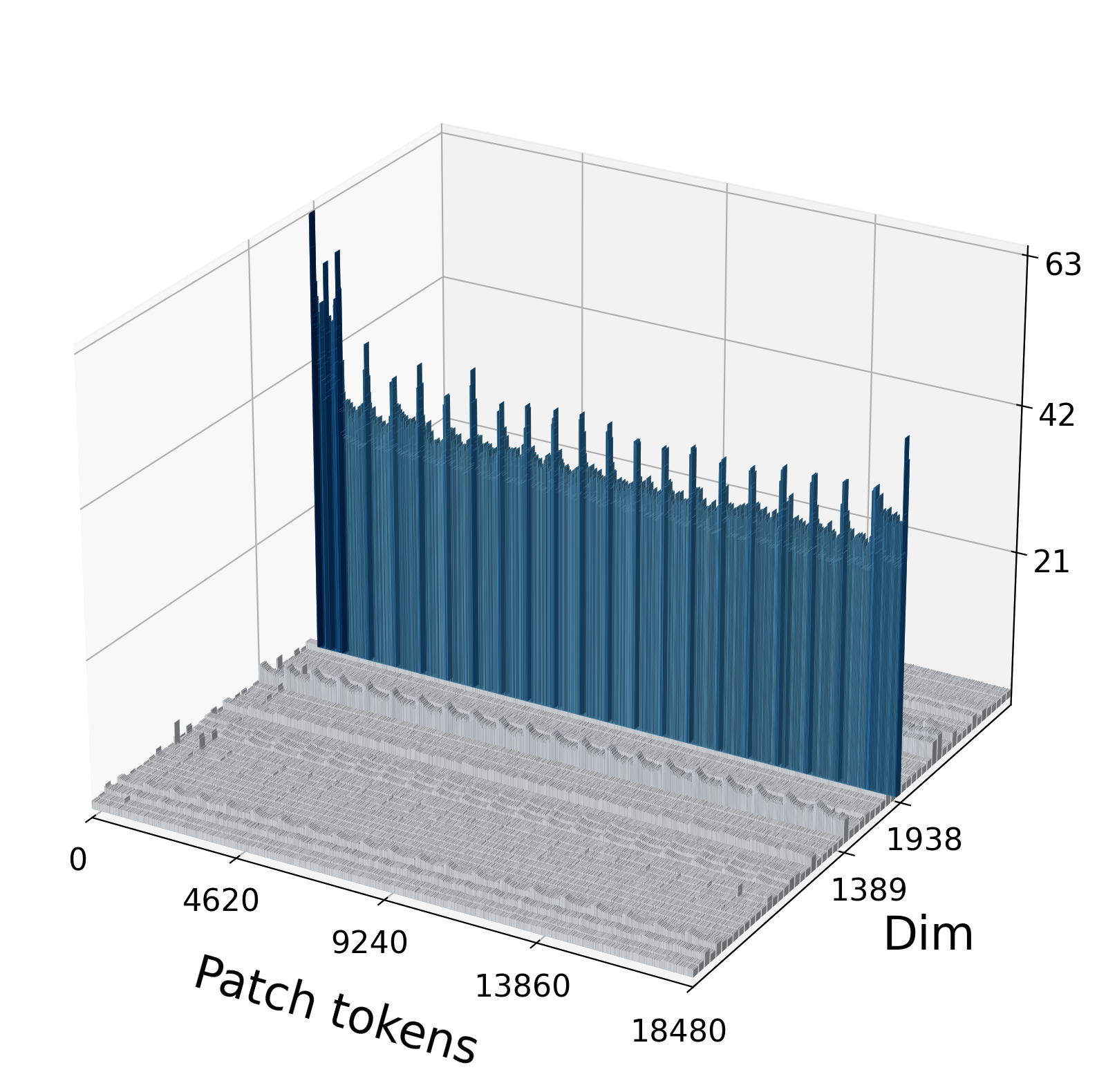}
    \caption{Wan2.2-5B}
    \label{fig:fig1b}
\end{subfigure}\hfill
\begin{subfigure}[t]{0.24\linewidth}
    \centering
    \includegraphics[width=\linewidth]{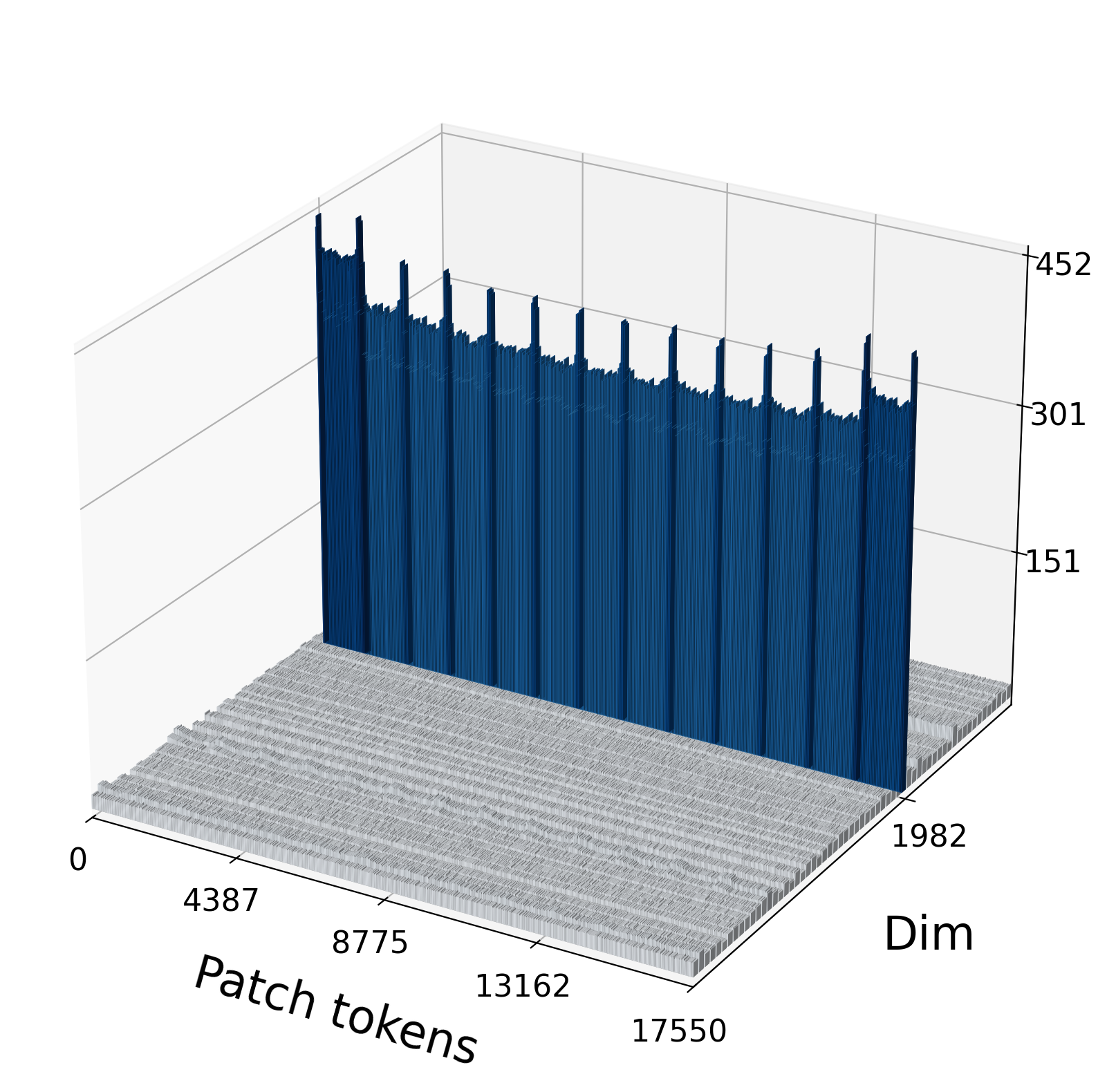}
    \caption{CogVideoX-5B}
    \label{fig:fig1c}
\end{subfigure}\hfill
\begin{subfigure}[t]{0.24\linewidth}
    \centering
    \includegraphics[width=\linewidth]{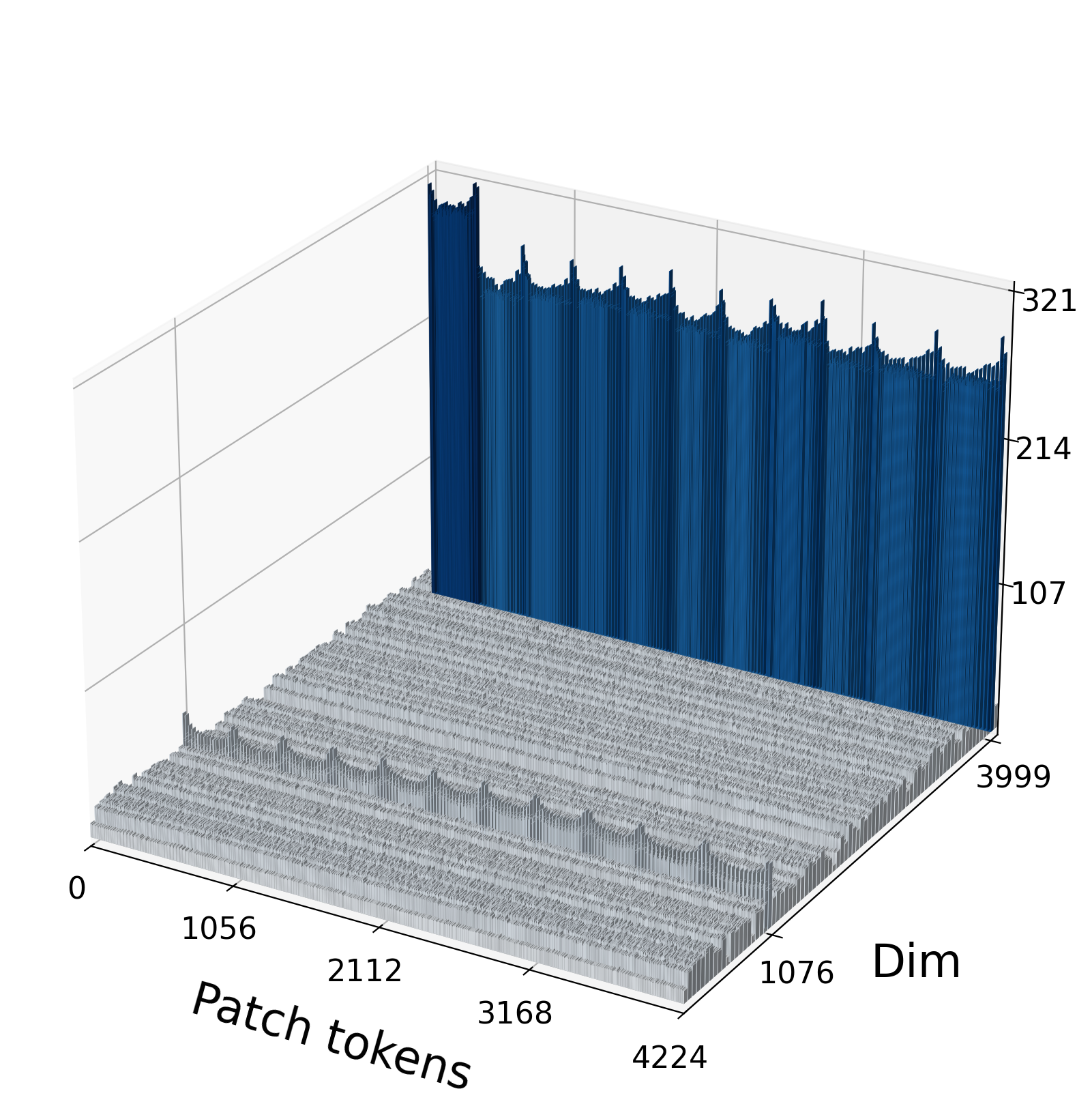}
    \caption{LTX2 (base)}
    \label{fig:fig1d}
\end{subfigure}


\caption{\textbf{Video DiTs exhibit a structured positional hierarchy of Massive
Activations.} Activation magnitudes across four video DiTs, averaged over 100
prompts and aggregated in bins of 60 consecutive tokens for visualization.
Across all models, MA magnitudes peak at tokens of the first latent frame and
recur at the spatial boundaries of subsequent latent frames.}

\label{fig:fig1}

\end{figure}

Taking a closer look at the internal activations of video DiTs, we find that MAs not only emerge, often exceeding $50{\times}$ the mean activation magnitude, but also exhibit a structured positional hierarchy (Figure~\ref{fig:fig1}). MA magnitudes are highest at tokens of the first latent frame, elevated at the spatial boundaries of subsequent latent frames, and comparatively moderate in their interiors, producing recurring peaks along the flattened token sequence. This pattern is consistent across Wan2.1-1.3B and Wan2.2-5B \cite{wan}, CogVideoX \cite{cogvideox}, and LTX-2 \cite{ltx2}, spanning both standard DiT and MM-DiT designs as well as different latent compression ratios of $4{\times}8{\times}8$, $4{\times}16{\times}16$, and $8{\times}32{\times}32$.

\begin{figure}[t]
    \centering
    \begin{subfigure}[b]{0.31\linewidth}
        \centering
        \includegraphics[width=\linewidth]{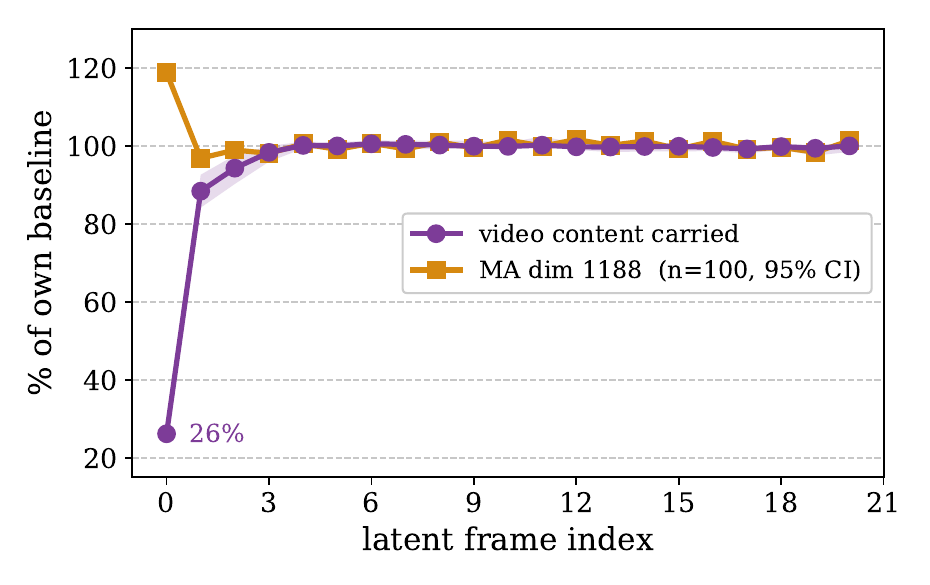}
        \caption{Per-frame content load vs.\ MA}
        \label{fig:vae_border_a}
    \end{subfigure}
    \hfill
    \begin{subfigure}[b]{0.675\linewidth}
        \centering
        \includegraphics[width=\linewidth]{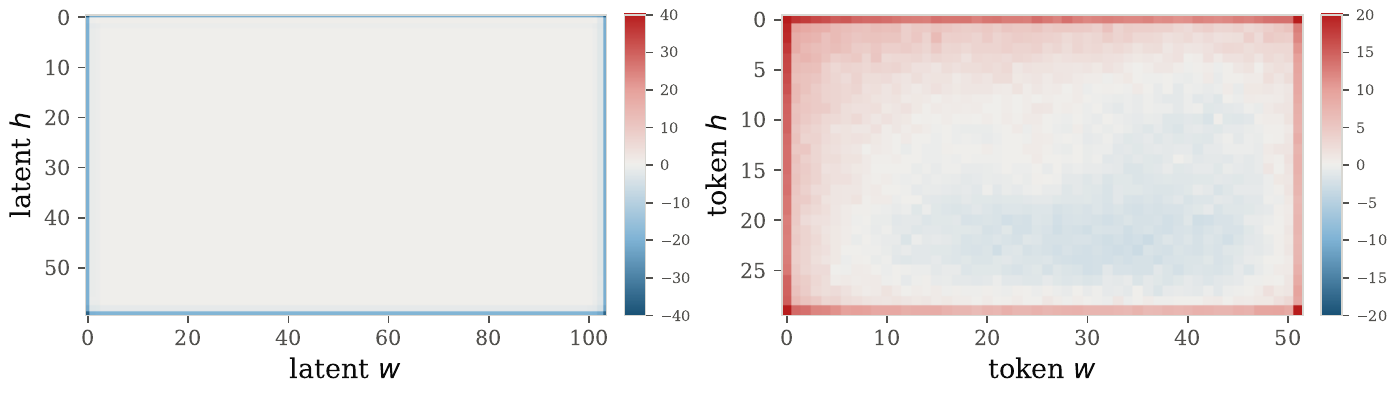}
        \caption{Latent boundaries of the VAE (left) and the DiT MA map (right)}
        \label{fig:vae_border_b}
    \end{subfigure}

  \caption{\textbf{The MA hierarchy aligns with video-VAE encoding asymmetries.}
(a)~Latent-frame ablation in Wan2.1 VAE shows that causal temporal padding leaves the first
latent frame carrying less content than later frames, precisely where MA
magnitudes peak. (b)~A uniform video isolates spatial padding: the VAE latent
has attenuated boundaries (left), and the DiT MA map (dimension 1188) is
elevated at the corresponding boundary tokens (right; shown at token
resolution, where each token aggregates a $2{\times}2$ patch of latent cells).}
    \label{fig:vae_border}
\end{figure}

We trace this positional hierarchy to the encoding asymmetries of the video VAE: causal temporal encoding leaves the first latent frame with only a fraction of the content load of a later frame, and zero spatial padding depresses frame borders, exactly where MAs are most elevated (Figure~\ref{fig:vae_border}). We then probe the computational role of MAs with bidirectional interventions and find that raising MA magnitude weakens the corresponding self-attention and feed-forward residual updates, whereas zeroing it strengthens them. Read together, these findings expose a native behavior: video DiTs use MAs to damp residual writing precisely where their encoder supplies the least content. This echoes the broader tendency of transformers to place sinks on less informative tokens, such as delimiters in LLMs and background patches in ViTs \cite{sun2024massive, darcet2024vision}. In video DiTs, however, these positions are structural, fixed by the encoder's geometry rather than by content.

Motivated by this native rescaling behavior, we propose \textbf{St}ructured \textbf{A}ctivation \textbf{S}teering (\textsc{\textbf{STAS}}), a training-free technique that strengthens it, steering MAs at first-frame and boundary positions toward a scaled reference derived from the model's own activations during early denoising. STAS modifies no model parameters and requires no additional model evaluations. It involves only element-wise assignments at selected token positions, introducing negligible overhead of less than $0.1\%$. Across different baselines, \textsc{STAS} consistently improves video quality and temporal coherence.

Our contributions are summarized as follows:
\begin{itemize}
    \item We systematically characterize MAs in video DiTs, revealing a positional hierarchy across token locations that is consistent across architectures and latent compression ratios, and we trace its origin to encoding asymmetries introduced by video VAEs.
    \item Through bidirectional interventions, we show that MA magnitude acts as an implicit rescaler of self-attention and feed-forward residual updates, with its strongest effects concentrated at the lower-content structural positions.
    \item We propose \textsc{STAS}, a training-free technique that exploits this structured rescaling signal, and demonstrate consistent improvements in temporal coherence and video quality across multiple text-to-video models, benchmark evaluations, and human preference studies, with negligible overhead and compatibility with other training-free methods.
\end{itemize}

\section{Related Work}


\textbf{Massive Activations.} Massive activations have been reported across transformer families spanning language \cite{sun2024massive, owen2025refined}, vision \cite{darcet2024vision}, and diffusion models \cite{massive, unleashing, trdq, tinyfusion}. In Large Language Models (LLMs), such activations often emerge at fixed dimensions and are frequently associated with low-information tokens, and have been linked to contextual knowledge modeling in several analyses \cite{jin2025massive}. Similar patterns are observed in Vision Transformers (ViTs), where MAs are reported to arise in redundant or background-like tokens and correlate with global semantic encoding \cite{darcet2024vision,yang2024denoising}. In DiTs, MAs have been discussed mainly due to their practical implications for acceleration techniques such as quantization \cite{trdq, hqdit, zhao2025viditq} and distillation \cite{tinyfusion}, and their presence can also affect extracted feature quality \cite{unleashing}. Recent studies further examine their generative role in image DiTs, relating them to local-detail synthesis or prompt-conditioned semantic subspaces \cite{massive, turri2026few}. However, how MA magnitudes are organized over spatiotemporal video tokens, and how they modulate residual computation across these positions, remain unexplored.

\textbf{Improving Video Generation with Training-Free Methods.} Prior works \cite{bytheway, unictrl} improve temporal consistency in U-Net-based video generation models by manipulating attention maps. Others use external guidance from image generators \cite{i4vgen, jagpal2025eidt} or multimodal LLMs \cite{li2025training}. FreeInit \cite{freeinit} and FreqPrior \cite{freqprior} refine noise priors to reduce spatiotemporal incoherence. Beyond external signals, some methods use the generator itself for guidance. Auto-guidance contrasts the original model with a weaker variant \cite{karras2024guiding}, obtained by skipping selected spatiotemporal layers \cite{hyung2025spatiotemporal} or dropping DiT blocks \cite{s2}. FlowMo \cite{flowmo} and TeDiO \cite{tursynbek2026tedio} optimize intermediate latents using temporal variation and self-attention diagonals, respectively; SRVS \cite{jang2026selfrefining} performs iterative predict-and-perturb updates, using prediction inconsistency to localize uncertain regions; and ANSE \cite{kim2025model} uses attention-derived uncertainty to select initial noise. All, however, typically either require additional model evaluations or impose operations the model does not natively perform. Our method instead strengthens a regulation the model already implements, the MA-based rescaling of its least informative tokens, within existing forward passes and with negligible overhead.

\section{Preliminaries: Video VAEs and DiTs}
\label{preliminaries}

Modern video DiTs generate in the latent space of a spatiotemporal VAE. An input video is encoded causally: the first physical frame forms the first latent frame on its own, and every subsequent chunk of $r_{\mathrm{temp}}$ frames forms one more latent frame. The encoder pads asymmetrically, front-only in time and symmetric zeros in space. Consequently, the receptive fields of the first latent frame and of the outermost spatial positions contain more padding and less video content than those of later frames and interior positions. The latent is patchified ($2{\times}2$ spatially) into video tokens and denoised by a stack of transformer blocks conditioned on the text prompt and timesteps, trained under the flow-matching objective \cite{flowmatching}. Finally, sampling proceeds iteratively over denoising steps $k = 0, 1, \dots$, from pure noise to the clean latent.

\section{The Organization and Function of MAs in Video DiTs}
\label{sec:observations}
Video DiTs inherit the MA properties reported for image DiTs \cite{massive}: MAs persist across layers and model scales, concentrate on fixed channel dimensions, and remain largely invariant to text prompts (Figure~\ref{fig:fig1}). Beyond these, we study MAs along the token axis, answering the two questions posed in Sec. \ref{sec:intro}: how MA magnitudes are organized over the video-token layout, and what function they serve during generation.

\begin{wrapfigure}{r}{0.42\linewidth}
    \vspace{-12pt}
    \centering
    \includegraphics[width=\linewidth]{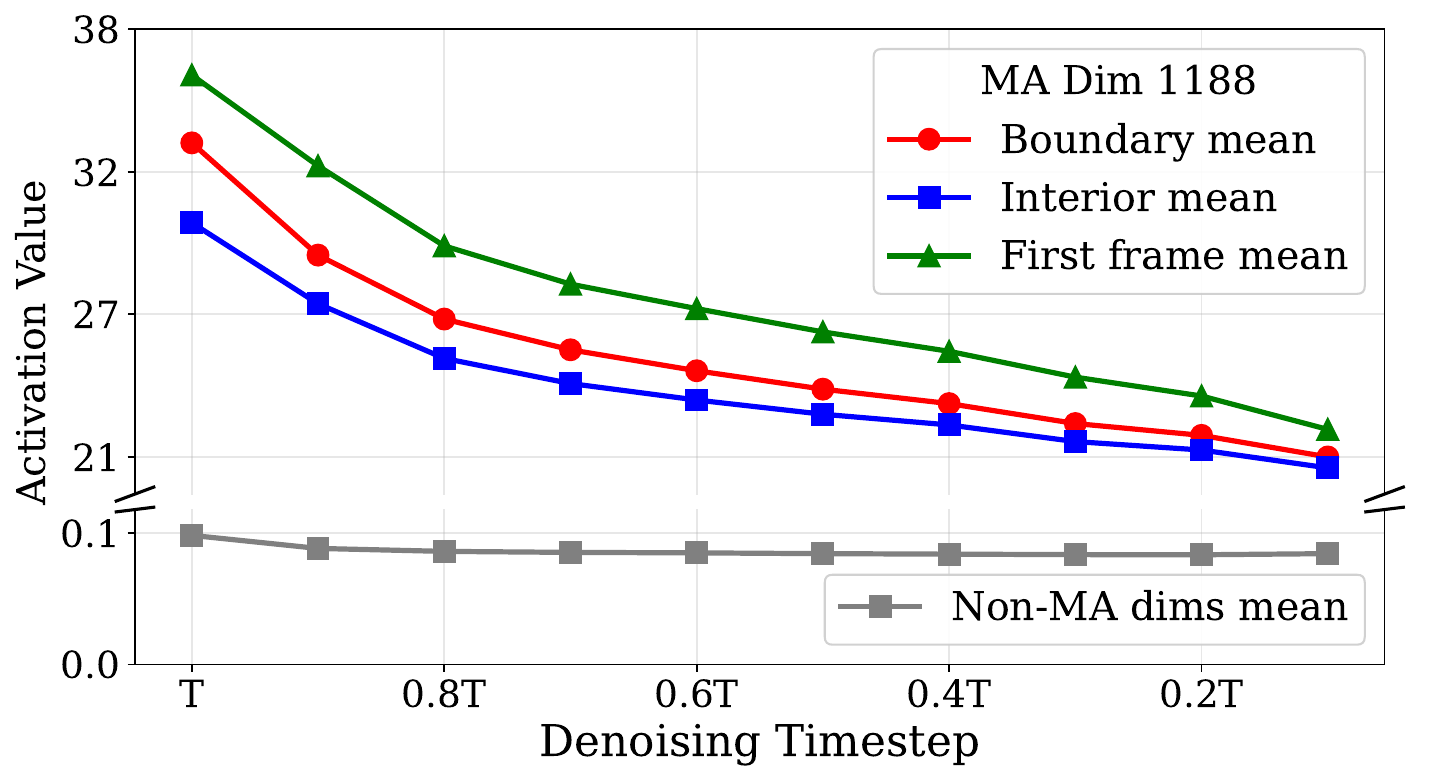}
    \caption{\textbf{The MA hierarchy is most pronounced during early denoising.}}
    \label{fig:decay}
    \vspace{-8pt}
\end{wrapfigure}


    

\paragraph{Periodic MA Pattern Emerges along the Token Sequence.} 
\label{obs:obs1}
We record hidden states during generation over 100 prompts and examine MA magnitudes along the flattened token sequence (Figure~\ref{fig:fig1}, see Appendix~\ref{app:ma_measurement} for the
measurement protocol). Each model typically exhibits a single dominant MA dimension. Their MA magnitudes follow a three-level hierarchy: highest at tokens of the first latent frame, elevated along the spatial boundaries of every subsequent latent frame, and moderate in frame interiors. Under row-major flattening, these boundaries appear as periodic head and tail bands. The hierarchy is robust across different video DiT designs and compression ratios.


The hierarchy has a characteristic timing: it is most pronounced during early denoising (Figure~\ref{fig:decay}), the phase deciding global structure, and attenuates monotonically as sampling progresses, with the boundary-to-interior ratio falling $1.10\!\to\!1.02$. This motivates the early-step window of our method.

\paragraph{The Hierarchy Mirrors the VAE's Encoding Asymmetry.}
\label{obs:obs2}
Why these positions? The padding facts from Sec. \ref{preliminaries} predict exactly where the encoder supplies less content, and two controlled probes confirm that the MA hierarchy lands on those positions.

Temporally, we measure the content each latent frame carries with an ablate-and-decode probe. We encode natural videos, replace one latent frame at a time with its spatial mean, decode, and measure the reconstruction damage on the physical frames that this latent frame generates (formalized in Appendix~\ref{app:experimental_protocal}). The first latent frame, which encodes a single physical frame, carries only about a quarter of a later frame's content load, and MA magnitudes peak exactly there (Figure~\ref{fig:vae_border_a}).


Spatially, we encode a video with no spatial structure, every frame a single constant color. The resulting latent still exhibits a depressed border ring (Figure~\ref{fig:vae_border_b}, left). Nothing in the input can cause it, so the ring must be the footprint of zero padding. The DiT MA map mirrors this ring (Figure~\ref{fig:vae_border_b}, right), and the correspondence is graded rather than binary: the four corners, padded twice over, show roughly twice the edge deficit and more than double the edge-level MA elevation. The same additivity holds across probes: border tokens of the first latent frame carry both shortfalls, and show the highest MA magnitudes in the sequence (Figure~\ref{fig:fig1}). Natural videos and re-encoded generations reproduce the ring (Appendix~\ref{app:additional_vae_ma}), showing that this prior is present in the model's own outputs.



The hierarchy thus inherits the geometry of the video VAE encoder, concentrating MAs at positions with reduced content load. This echoes a broader tendency of transformers to offload onto less informative tokens, previously described as attention sinks~\cite{sun2024massive, darcet2024vision}. Here it appears instead as a magnitude hierarchy on a single MA dimension, at positions fixed by the encoder rather than by the content of each input. The model locates them through RoPE~\cite{rope}, and permuting the temporal or spatial indices relocates the MA peaks (Appendix~\ref{app:rope}).

\paragraph{MA Magnitude Acts as a Token-Level Rescaler of Residual Writing.}
\label{obs:obs3}
What do these elevated MAs do? In pre-normalized DiT blocks, a dominant MA occupies a disproportionate share of a token's normalized scale, attenuating the remaining content-bearing dimensions, consistent with the analysis of residual sinks \cite{qiu2026unified}. This attenuation directly enters the feed-forward and attention-value pathways, whereas query/key normalization reduces scale variation but not directional changes. We give further discussions in Appendix~\ref{app:rescaler_math} and hypothesize a two-fold rescaling effect: increasing a token's MA directly weakens its own residual writing and, through self-attention, can indirectly alter the attention updates received by other video tokens. In contrast, token-local computation at non-target positions and the unmodified text pathway should be less affected.

\begin{wrapfigure}{r}{0.42\linewidth}
    \vspace{-16pt}
    \centering
    \includegraphics[width=\linewidth]{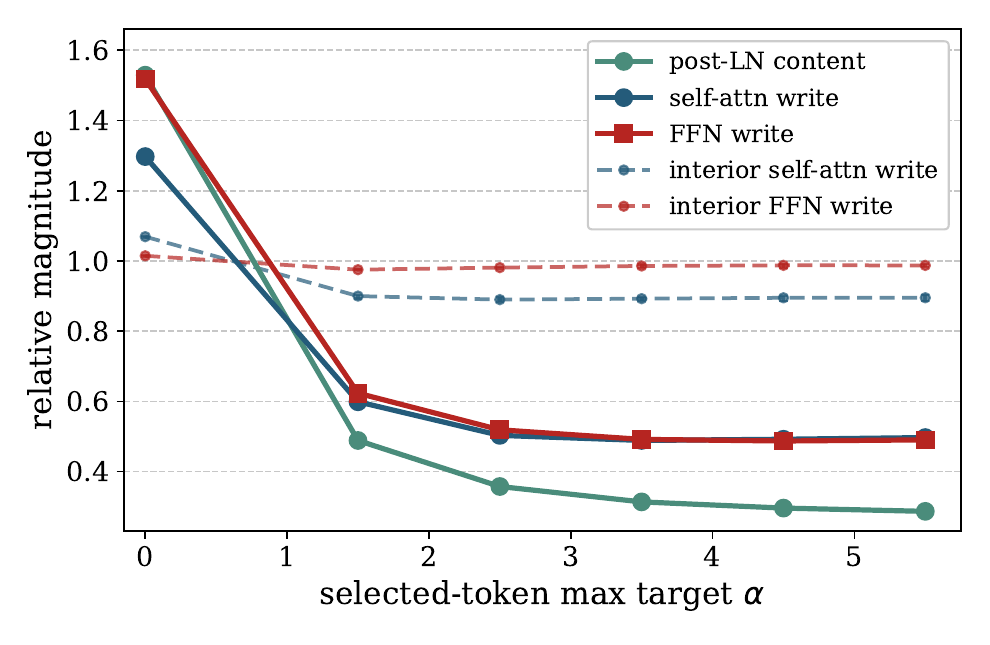}
    \caption{\textbf{MA magnitude is a token-level rescaler of residual updates.}}
    \label{fig:rescale}
    \vspace{-14pt}
\end{wrapfigure}

We test this hypothesis on Wan2.1-1.3B by steering the MA dimension at structural tokens, the first latent frame and the head-and-tail bands of later frames, which cover the contiguous sides of the border ring under flattening, toward $\alpha$ times its global maximum, sweeping $\alpha$ from $0$ (erasure) to $5.5$ over 100 paired generations (setup in Appendix \ref{app:rescale}). In the following block, we measure post-normalization content and the self-attention and feed-forward residual writes. As shown in Figure~\ref{fig:rescale}, the structural tokens respond monotonically in both directions: MA erasure amplifies their writes to $1.3$--$1.5\times$ the native values, whereas $\alpha{=}2.5$ reduces their self-attention and feed-forward writes to $0.50\times$ and $0.52\times$, respectively. The main non-target response appears in interior self-attention writes ($1.07\times$ under erasure and $0.89\times$ at $\alpha{=}2.5$), consistent with propagation through token mixing, while interior feed-forward writes remain near native. These results identify MA magnitude as a token-level rescaler of both self-attention and feed-forward writes, whose reach to other tokens runs through self-attention.

\section{Method}
\label{sec:method}

The analysis in Sec.~\ref{sec:observations} reveals that MA magnitude acts as an implicit, token-level rescaler of residual computation, and that video DiTs deploy it unevenly, concentrating elevated MAs at VAE-aligned structural positions with reduced content load. We hypothesize that residual writing at these positions, when insufficiently damped, injects perturbations during early denoising that propagate through self-attention and persist along the denoising trajectory, surfacing as degraded first frames, incoherent motion, and abrupt object appearances and disappearances. This motivates us to strengthen the model's native rescaling at its existing structural locations.

We therefore propose \textsc{STAS}, a training-free method that specifies \emph{what} to steer (MA dimensions), \emph{where} to steer (VAE-aligned structural tokens), \emph{how} to steer them (max-referenced calibration), and \emph{when} to intervene (early denoising). Let $H^{m}_{k, i, d}$ denote the hidden dimension $d$ of token $i$ after DiT block $m$ at sampling step $k$, where $k=0$ denotes the first denoising timestep. STAS modifies only the model-specific MA dimensions $\mathcal{M}$ identified by activation profiling. Its target token set is
\begin{equation}
  \mathcal{S}=\mathcal{F}_0\cup\mathcal{B}(p),
  \label{eq:stas_S}
\end{equation}
where $\mathcal{F}_0$ contains all tokens in the first latent frame and $\mathcal{B}(p)$ contains the head and tail $p\%$ of the flattened tokens within each latent frame. 
In our default implementation, $\mathcal{B}(p)$ focuses on the horizontal boundary bands for two complementary reasons. Under row-major flattening, the head and tail rows form contiguous spans that capture the prominent recurring MA peaks, whereas the left and right columns are sparse and non-contiguous. Moreover, the tail band of one latent frame and the head band of the next lie on opposite sides of each latent frame-frame junction, making them a compact, ordering-aligned choice for steering around frame transitions.

\begin{wrapfigure}{r}{0.38\linewidth}
\vspace{-28pt}
\centering
\includegraphics[width=\linewidth]{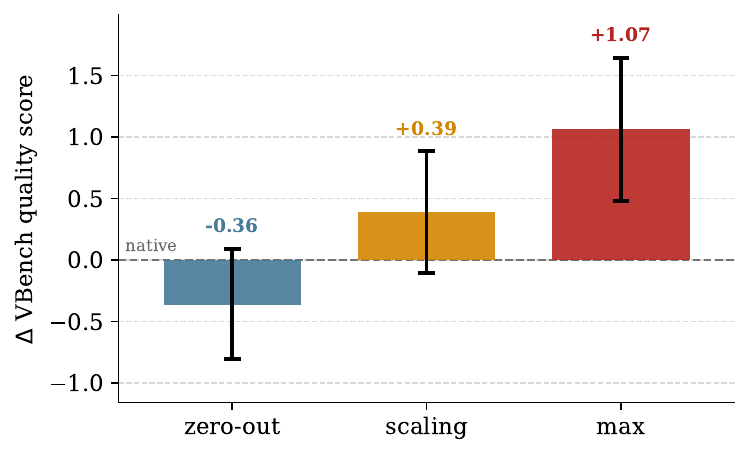}
\caption{\textbf{Steering direction and calibration rule both matter.}}
\label{fig:calibration_comparison}
\vspace{-14pt}
\end{wrapfigure}

Although structural positions exhibit higher MAs on average, their native magnitudes remain heterogeneous: some still fall within the interior-token range (as detailed in Appendix~\ref{app:coverage}). Multiplying
each token by a fixed factor preserves this heterogeneity and may leave weak structural MAs comparatively small. STAS instead derives an adaptive reference from the current hidden state:
\begin{equation}
  r^{m}_{k,d}
  =
  \max_{j\in\mathcal{V}}
  \left|H^{m}_{k,j,d}\right|,
  \label{eq:stas_reference}
\end{equation}
where $\mathcal{V}$ denotes the video tokens. It then defines the steering target as
\begin{equation}
g^{m}_{k,i,d}
=
\alpha \cdot r^{m}_{k,d}
\cdot \operatorname{sign}\!\left(H^{m}_{k,i,d}\right),
\label{eq:stas_target}
\end{equation}

where $\alpha$ controls the steering strength. This max-referenced rule gives selected tokens a consistent target magnitude while adapting to each sample, layer, and denoising step, with their native
polarity preserved. As shown in Figure~\ref{fig:calibration_comparison}, MA erasure (zeroing out) produces a negative quality trend and fixed scaling yields only a modest improvement, whereas max-referenced
calibration produces the clearest positive response.

STAS is applied only during the first $K$ denoising iterations, when the structural MA hierarchy is most pronounced and residual perturbations have more opportunity to affect the subsequent trajectory.
Let $\mathcal{L}$ denote the selected DiT blocks. The complete update is
  
\begin{equation}
    \widehat{H}^{m}_{k,i,d} =
    \begin{cases}
    g^{m}_{k,i,d}, & k<K,\ m\in\mathcal{L},\ i\in\mathcal{S},\ d\in\mathcal{M},\\[2pt]
    H^{m}_{k,i,d}, & \text{otherwise},
    \end{cases}
    \label{eq:stas}
\end{equation}
Eligible entries are therefore replaced by their calibrated targets, while all other activations follow the original denoiser unchanged. We provide the full algorithm in Appendix \ref{app:algorithm}. We also discuss our method through the lens of guidance in Appendix \ref{app:guidance_view}.

In practice, we apply \textsc{STAS} on top of CFG \cite{cfg}. By calibrating MA dimensions at the structural token positions (first-frame and boundary tokens), \textsc{STAS} complements CFG's global, prompt-level guidance with token-level regulation of residual writing.

\section{Experiments}
\label{sec:experiments}

We evaluate \textsc{STAS} on four publicly available text-to-video models spanning different architectures and scales: Wan2.1-T2V-1.3B~\cite{wan}, Wan2.2-TI2V-5B~\cite{wan}, CogVideoX-5B~\cite{cogvideox}, and LTX-2 base model ~\cite{ltx2}. All experiments use the officially released weights and default sampling configurations; unless otherwise specified, \textsc{STAS} is applied on top of the default CFG. Videos are generated at $81{\times}480{\times}832$ for Wan2.1, $49{\times}480{\times}720$ for CogVideoX-5B, $81{\times}704{\times}1280$ for Wan2.2, and $81{\times}512{\times}768$ for LTX-2. Implementation details are deferred to the Appendix \ref{app:model_configs}.

\paragraph{Evaluation Metrics.}
Our primary benchmark is VBench~\cite{vbench}, which aggregates quality (seven dimensions) and semantic (nine dimensions) scores into a total score on a 0--100 scale. All VBench results are averaged over five random seeds, over 4,700 videos per model and setting. Since VBench emphasizes per-dimension appearance and consistency, we additionally report the Fréchet Video Motion Distance (FVMD)~\cite{fvmd}, which compares motion-feature statistics between generated and real videos.

\paragraph{Quantitative Comparisons.}
As shown in Table~\ref{tab:quality_scores}, \textsc{STAS} consistently improves the aggregate VBench scores across all four backbones. The improvements generalize from standard DiT architectures, including Wan2.1 and Wan2.2, to the MM-DiT-based CogVideoX and the more aggressively compressed LTX-2, demonstrating broad applicability across architectures, model scales, and latent geometries. Notably, on every backbone, the semantic score improves at least as much as the quality score, consistent with our finding that \textsc{STAS} regulates video-token residual writing while affecting text-conditioned computation less directly. \textsc{STAS} also preserves or improves every visual-quality dimension in nearly all settings, without training or additional denoiser evaluations. On LTX-2, the gains concentrate in semantic alignment, possibly because its baseline already exhibits strong temporal stability, leaving less headroom for further
quality improvements. Nevertheless, the improved semantic and aggregate scores show that \textsc{STAS} remains effective under this distinct latent geometry. 

Following~\cite{kim2025model}, we further evaluate motion fidelity using combined FVMD~\cite{fvmd} on the MSR-VTT test set~\cite{msrvtt} in Table~\ref{tab:fvmd}. FVMD represents tracked keypoint
trajectories using velocity and acceleration descriptors. Its combined variant concatenates these first- and second-order motion features before computing the Fréchet distance between generated and
real-video distributions, with lower values indicating better motion fidelity. We report the mean and 95\% confidence interval over ten paired 200-video subsets. Our method consistently reduces FVMD across
all four backbones. These results complement VBench with distribution-level evidence of improved motion
fidelity across architectures and latent geometries.

\newsavebox{\vbenchbox}
\begin{table*}[t]
\centering
\begingroup

\captionsetup{
    font=small,
    skip=3pt,
    belowskip=0pt
}

\sbox{\vbenchbox}{%
\begin{minipage}[t]{0.76\textwidth}
\vspace{0pt}
\centering

\caption{\textbf{VBench results across different video backbones.}}
\label{tab:quality_scores}

\begingroup
\scriptsize
\setlength{\tabcolsep}{2.6pt}
\renewcommand{\arraystretch}{0.90}

\resizebox{\linewidth}{!}{%
\begin{tabular}{@{}llccccccc@{}}
\toprule
Backbone
& Method
& \makecell{Subject\\Cons.$\uparrow$}
& \makecell{Background\\Cons.$\uparrow$}
& \makecell{Aesthetic\\Quality$\uparrow$}
& \makecell{Imaging\\Quality$\uparrow$}
& \makecell{Quality\\Score$\uparrow$}
& \makecell{Semantic\\Score$\uparrow$}
& \makecell{Total\\Score$\uparrow$} \\
\midrule

\multirow{2}{*}{Wan2.1-1.3B}
& Vanilla
& 94.63 & 95.81 & 61.91 & 68.14 & 81.81 & 79.70 & 81.39 \\
& \cellcolor{orange!15}+Ours
& \cellcolor{orange!15}\textbf{95.00}
& \cellcolor{orange!15}\textbf{95.93}
& \cellcolor{orange!15}\textbf{62.03}
& \cellcolor{orange!15}\textbf{68.95}
& \cellcolor{orange!15}\textbf{82.03}
& \cellcolor{orange!15}\textbf{80.66}
& \cellcolor{orange!15}\textbf{81.76} \\
\midrule

\multirow{2}{*}{CogVideoX-5B}
& Vanilla
& 93.40 & 95.29 & 59.98 & 64.62 & 79.78 & 77.59 & 79.34 \\
& \cellcolor{orange!15}+Ours
& \cellcolor{orange!15}\textbf{93.80}
& \cellcolor{orange!15}\textbf{95.47}
& \cellcolor{orange!15}\textbf{60.31}
& \cellcolor{orange!15}\textbf{65.12}
& \cellcolor{orange!15}\textbf{79.95}
& \cellcolor{orange!15}\textbf{78.24}
& \cellcolor{orange!15}\textbf{79.61} \\
\midrule

\multirow{2}{*}{Wan2.2-5B}
& Vanilla
& 95.13 & 96.63 & 61.67 & 69.02 & 81.75 & 81.68 & 81.74 \\
& \cellcolor{orange!15}+Ours
& \cellcolor{orange!15}\textbf{95.37}
& \cellcolor{orange!15}\textbf{96.70}
& \cellcolor{orange!15}\textbf{61.72}
& \cellcolor{orange!15}\textbf{69.39}
& \cellcolor{orange!15}\textbf{81.82}
& \cellcolor{orange!15}\textbf{82.35}
& \cellcolor{orange!15}\textbf{81.93} \\
\midrule

\multirow{2}{*}{LTX-2 (base)}
& Vanilla
& 92.03 & 94.73 & \textbf{60.34} & \textbf{66.36}
& 81.11 & 76.83 & 80.26 \\
& \cellcolor{orange!15}+Ours
& \cellcolor{orange!15}\textbf{92.11}
& \cellcolor{orange!15}\textbf{94.77}
& \cellcolor{orange!15}\textbf{60.34}
& \cellcolor{orange!15}66.35
& \cellcolor{orange!15}\textbf{81.13}
& \cellcolor{orange!15}\textbf{77.32}
& \cellcolor{orange!15}\textbf{80.37} \\
\bottomrule
\end{tabular}%
}
\endgroup
\end{minipage}%
}

\begin{minipage}[t]{0.76\textwidth}
\vspace{0pt}
\usebox{\vbenchbox}
\end{minipage}
\hfill
\begin{minipage}
    [t]
    [\dimexpr\ht\vbenchbox+\dp\vbenchbox\relax]
    [s]
    {0.21\textwidth}
\vspace{0pt}
\centering

\includegraphics[
    width=\linewidth,
    trim=2pt 2pt 2pt 2pt,
    clip
]{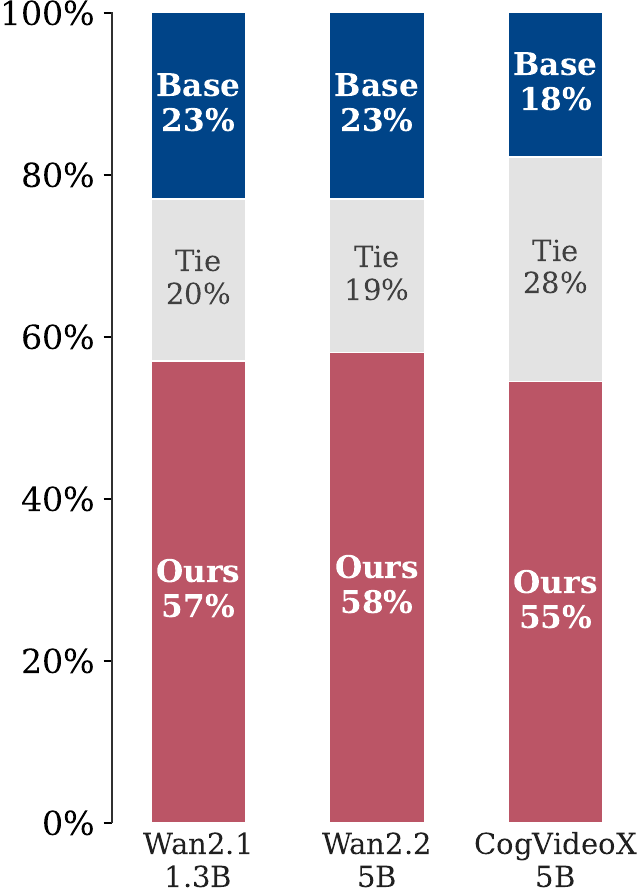}

\vfill

\captionsetup{
    type=figure,
    font=small,
    skip=3pt,
    belowskip=0pt
}
\captionof{figure}{\textbf{User study.}}
\label{fig:user_study}

\end{minipage}

\endgroup
\end{table*}

\begin{wraptable}{r}{0.49\linewidth}
\vspace{-6pt}
\centering
\scriptsize
\setlength{\tabcolsep}{2.4pt}
\renewcommand{\arraystretch}{0.92}

\caption{\textbf{FVMD metric comparisons.}}
\label{tab:fvmd}
\vspace{-4pt}

\begin{tabular}{@{}llc@{}}
\toprule
Backbone & Method & FVMD$\downarrow$ [95\% CI] \\
\midrule

\multirow{2}{*}{Wan2.1-1.3B}
& Vanilla & 11111.8 [10578.6, 11645.0] \\
& \cellcolor{orange!15}+Ours
& \cellcolor{orange!15}
  \textbf{10775.8 [10293.5, 11258.1]} \\
\midrule

\multirow{2}{*}{Wan2.2-5B}
& Vanilla & 11524.0 [11083.8, 11964.2] \\
& \cellcolor{orange!15}+Ours
& \cellcolor{orange!15}
  \textbf{11366.6 [10865.9, 11867.3]} \\
\midrule

\multirow{2}{*}{CogVideoX-5B}
& Vanilla & 14647.2 [14027.3, 15267.1] \\
& \cellcolor{orange!15}+Ours
& \cellcolor{orange!15}
  \textbf{13291.3 [12873.9, 13708.7]} \\
\midrule

\multirow{2}{*}{LTX-2 (base)}
& Vanilla & 12586.0 [11906.9, 13265.1] \\
& \cellcolor{orange!15}+Ours
& \cellcolor{orange!15}
  \textbf{12481.1 [11840.0, 13122.2]} \\
\bottomrule
\end{tabular}

\vspace{7pt}

\caption{\textbf{Results on T2V-CompBench.}}
\label{tab:t2v_compbench}
\vspace{-4pt}

\setlength{\tabcolsep}{2.1pt}
\begin{tabular}{@{}llccc@{}}
\toprule
Backbone
& Method
& \makecell{Consistent\\Attr.$\uparrow$}
& \makecell{Dynamic\\Attr.$\uparrow$}
& \makecell{Time\\(s)$\downarrow$} \\
\midrule

\multirow{2}{*}{Wan2.1-1.3B}
& Vanilla
& 76.27 & 2.72 & 234.47 \\
& \cellcolor{orange!15}+Ours
& \cellcolor{orange!15}\textbf{77.36}
& \cellcolor{orange!15}\textbf{3.40}
& \cellcolor{orange!15}234.53 \\
\bottomrule
\end{tabular}

\vspace{-18pt}
\end{wraptable}

We additionally evaluate \textsc{STAS} on subsets of T2V-CompBench~\cite{t2v_compbench}, which assess object--attribute consistency under static and temporally changing conditions. Due to limited computational resources, we conduct this evaluation on Wan2.1-1.3B. As shown in Table~\ref{tab:t2v_compbench}, \textsc{STAS} improves both metrics while leaving inference time virtually unchanged, further supporting its effectiveness.


\paragraph{Computational Cost.} We report inference time on an NVIDIA RTX A6000 GPU, averaged over 5 runs after 100 warm-up denoising steps. As shown in Table \ref{tab:t2v_compbench}, our method increases runtime from 234.47\,s to 234.53\,s, indicating negligible inference-time overhead. 
This is expected since \textsc{STAS} introduces no additional forward computation and only applies an in-place update to a small set of MA dimensions (only one dimension in practice).

\paragraph{Adding \textsc{STAS} to Existing Training-Free Methods.}
\textsc{STAS} is orthogonal and complementary to existing training-free methods for improving video generation and can be added on top of them to boost performance. We demonstrate this by integrating \textsc{STAS} with
FlowMo~\cite{flowmo} and CFG-Zero$^*$~\cite{cfg0} on Wan2.1-1.3B, and report the quality dimensions in Table~\ref{tab:cfgzero_flowmo_ours}. Adding \textsc{STAS} improves most quality dimensions and increases
the overall quality score for both CFG-Zero$^*$ and FlowMo, with gains in subject/background consistency, temporal stability, and imaging quality (e.g., +0.99 and +0.78, respectively). While most
dimensions improve, we observe a mild drop in dynamic degree. This is expected because dynamic degree primarily measures motion magnitude rather than temporal coherence, and incoherent motion such as
flickering can artificially inflate this score. The net effect is a consistent improvement in generation quality, while the slight reduction in dynamic degree may be a natural byproduct of improved
coherence.

\begin{figure}[!t]                                                       
      \centering
      \includegraphics[width=0.995\textwidth]{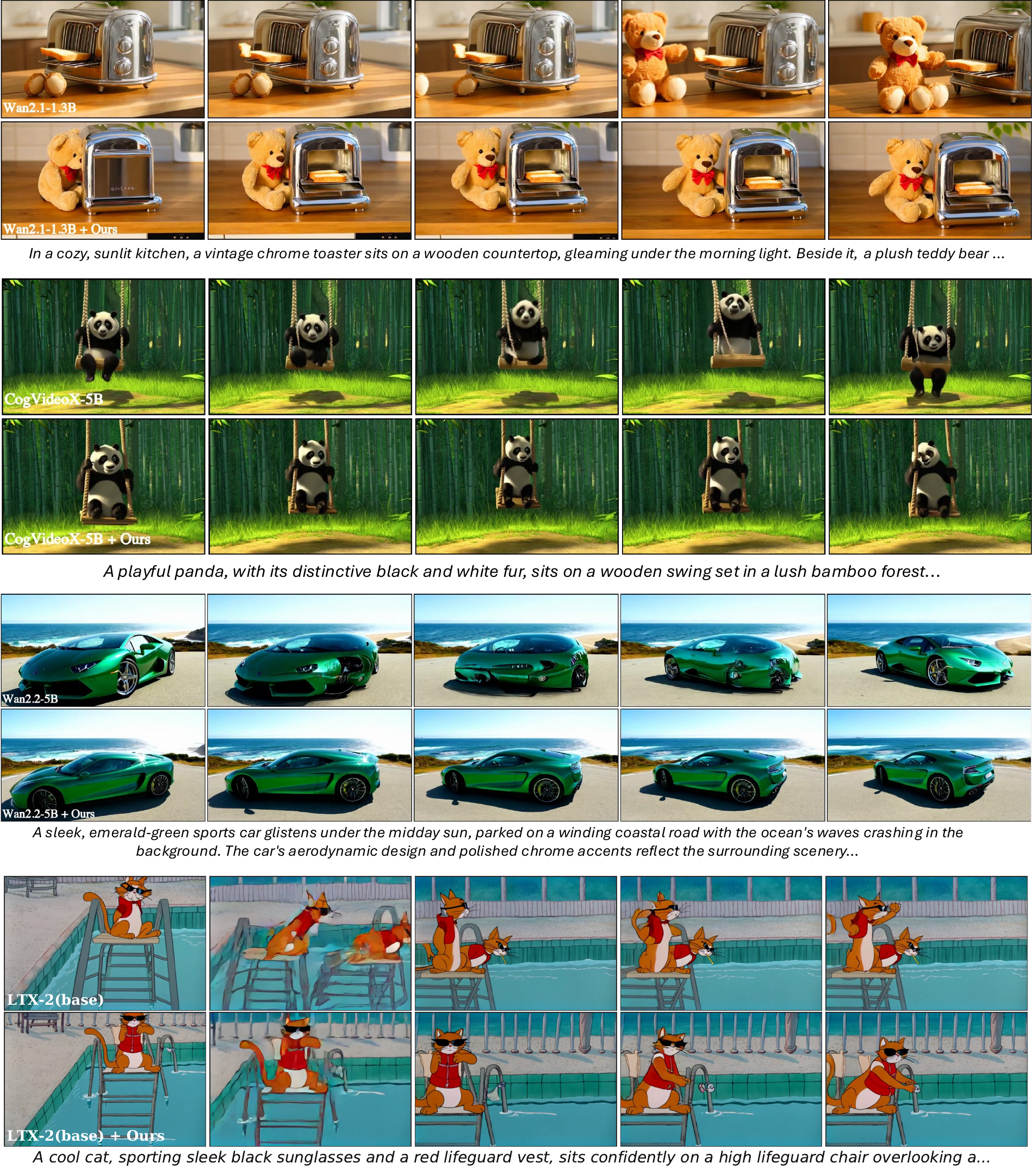}
      \vspace{-14pt}
      \caption{\textbf{Qualitative comparison with and without \textsc{STAS} on different backbones.}}
      \label{fig:fig5}
      \vspace{-14pt}
\end{figure}

\paragraph{Qualitative Comparisons.}
  We present qualitative comparisons across four backbones in
  Figure~\ref{fig:fig5}. \textsc{STAS} improves temporal coherence while preserving visual
  quality: it prevents abrupt teddy-bear appearance on Wan2.1, maintains panda
  identity on CogVideoX, preserves car geometry on Wan2.2, and suppresses cat
  duplication on LTX-2. We further conduct a blinded user study with nine
  participants and 20 randomized video pairs per model. \textsc{STAS} is preferred in
  57\%, 58\%, and 55\% of comparisons on Wan2.1, Wan2.2, and CogVideoX,
  respectively, consistently exceeding the corresponding baselines
  (Figure~\ref{fig:user_study}). Additional results are provided in the Appendix \ref{app:additional_results}.

\section{Ablation Study}
We mainly ablate four design choices of \textsc{STAS} on Wan2.1-1.3B model: \emph{what} to steer (MA dimension $\mathcal{M}$), \emph{where} to steer (target token set $\mathcal{S}$), \emph{when} to steer (timestep threshold $K$), and \emph{how} to steer (the calibration rule $g$). More ablations are provided in Appendix \ref{app:ablation}.


\begin{table}[t]
  \centering
  \caption{\textbf{Compatibility with CFG-Zero$^*$ and FlowMo.}}
  \label{tab:cfgzero_flowmo_ours}
  \scriptsize
  \setlength{\tabcolsep}{2.6pt}
  \renewcommand{\arraystretch}{0.90}
  \begin{tabular}{@{}lcccccccc@{}}
  \toprule
  Method
  & \makecell{Subject\\Cons.$\uparrow$}
  & \makecell{Background\\Cons.$\uparrow$}
  & \makecell{Temporal\\Flicker$\uparrow$}
  & \makecell{Motion\\Smooth.$\uparrow$}
  & \makecell{Dynamic\\Degree$\uparrow$}
  & \makecell{Aesthetic\\Quality$\uparrow$}
  & \makecell{Imaging\\Quality$\uparrow$}
  & \makecell{Quality\\Score$\uparrow$} \\
  \midrule
  CFG-Zero$^*$
  & 93.43 & 94.88 & 97.23 & 98.43
  & \textbf{58.41} & 62.17 & 67.03 & 81.81 \\
  \rowcolor{orange!15}
  + \textsc{STAS}
  & \textbf{93.93} & \textbf{95.09} & \textbf{97.42}
  & \textbf{98.54} & 57.69 & \textbf{62.35}
  & \textbf{68.02} & \textbf{82.21} \\
  \midrule
  FlowMo
  & 93.96 & 96.40 & 97.38 & 98.50
  & \textbf{56.57} & 65.51 & 67.90 & 82.83 \\
  \rowcolor{orange!15}
  + \textsc{STAS}
  & \textbf{94.12} & \textbf{96.42} & \textbf{97.51}
  & \textbf{98.55} & 55.78 & \textbf{65.53}
  & \textbf{68.68} & \textbf{83.00} \\
  \bottomrule
  \end{tabular}
  \end{table}

\begin{table}[t]
\centering
\scriptsize
\caption{
\textbf{Ablation study of \textsc{STAS} on Wan2.1-1.3B.}
}
\setlength{\tabcolsep}{2.5pt}
\begin{tabular}{lccccccccc}
\toprule
Variant & Dim & Target Tokens & $K$ & Rule
& \makecell{Sub.\\Consist.}
& \makecell{BG.\\Consist.}
& \makecell{Aesthetic\\Quality}
& \makecell{Imaging\\Quality}
& \makecell{Quality\\Score} \\
\midrule
Vanilla 
& -- & -- & -- & -- 
& 94.36 & 95.68 & 63.28 & 67.21 & 82.01 \\
\midrule
\rowcolor{gray!10}
\multicolumn{10}{l}{\textit{What (Dimension)}} \\
\quad Non-MA
& 42 & first+boundary & 20 & max
& 94.33 & 95.70 & 63.24 & 67.23 & 81.97 \\
\quad Weak-MA
& 71 & first+boundary & 20 & max
& 94.24 & 95.61 & 63.15 & 67.22 & 81.99 \\
\midrule
\rowcolor{gray!10}
\multicolumn{10}{l}{\textit{Where (Token Position)}} \\
\quad First 
& 1188 & first & 20 & max
& 94.25 & 95.52 & 63.24 & 67.58 & 82.04 \\
\quad Boundary 
& 1188 & boundary & 20 & max
& \textbf{94.79} & \textbf{95.84} & \underline{63.45} & 67.96 & 82.09 \\
\quad Interior 
& 1188 & first+interior & 20 & max
& 94.72 & 95.73 & 63.26 & 68.08 & 82.09 \\
\midrule
\rowcolor{gray!10}
\multicolumn{10}{l}{\textit{When (Timestep)}} \\
\quad $K=5$ 
& 1188 & first+boundary & 5 & max
& 94.68 & \underline{95.81} & 63.39 & 67.73 & \underline{82.13} \\
\quad $K=50$ 
& 1188 & first+boundary & 50 & max
& 94.62 & 95.73 & 63.31 & \textbf{68.13} & \underline{82.13} \\
\midrule
\rowcolor{gray!10}
\multicolumn{10}{l}{\textit{How (Amplification Rule)}} \\
\quad Scaling
& 1188 & first+boundary & 20 & scaling
& 94.51 & 95.71 & \textbf{63.47} & 67.56 & 82.06 \\
\midrule
\rowcolor{orange!15}
\textbf{Ours(\textsc{STAS})} 
& 1188 & first+boundary & 20 & max
& \underline{94.74} & \textbf{95.84} & \textbf{63.47} & \underline{68.05} & \textbf{82.24} \\
\bottomrule
\end{tabular}
\vspace{-8pt}
\label{tab:ablation}
\end{table}


\paragraph{Effect of dimension choice $\mathcal{M}$.} We first ablate the choice of the steered activation dimension. Table \ref{tab:ablation} shows that steering non-MA (dim 42, randomly selected, ${\sim}1\times$ peak-to-mean ratio) or weak-MA\footnote{ We term dimensions with moderate peak-to-mean ratios ($5$--$20\times$) ``weak-MA''.} (dim 71, ${\sim}6\times$) dimensions yields negligible or even negative changes, in contrast to the MA dimension (dim 1188, $59\times$). This confirms that meaningful activation spikes are essential for effective steering.
In contrast, all variants operating on the MA dimension achieve notable improvements.

\paragraph{Effect of steering boundary tokens $\mathcal{S}$.} Table~\ref{tab:ablation} compares different steering token positions. Boundary steering outperforms first-frame steering, while replacing boundary tokens with interior tokens lowers overall quality. Combining first-frame and boundary tokens performs best.

\vspace{-2pt}
\paragraph{Effect of timestep threshold $K$.}
We further ablate the timestep threshold $K$ for applying \textsc{STAS}. As shown in Table \ref{tab:ablation}, all tested settings improve over the vanilla baseline, while $K{=}20$ yields the best overall performance. In comparison, a smaller $K$ (e.g., $K{=}5$) provides weaker gains, and a larger $K$ (e.g., $K{=}50$) becomes less balanced across metrics. This matches our observation in Figure \ref{fig:decay} that MA magnitudes decay over denoising, suggesting that steering is most beneficial during the early high-noise steps when global structure is formed.

\vspace{-2pt}
\paragraph{Effect of amplification rule $g$.}
We compare max-referenced calibration with fixed scaling. Scaling preserves the heterogeneous magnitudes of structural MAs, leaving weak activations relatively small. By instead calibrating selected
tokens to the current peak activation magnitude while preserving their signs, our max-based rule achieves the best overall quality in Table~\ref{tab:ablation}.

\section{Limitations}
\textsc{STAS} strengthens the model's native suppression of residual writing at selected structural positions, and attention propagates this to other tokens, which may slightly attenuate subtle motion. Samples near the decision boundary can then fall below threshold-based metrics such as Dynamic Degree and be counted as static; the same trade-off is reported for other training-free methods~\cite{flowmo, jang2026selfrefining, tursynbek2026tedio} and even for post-training ones~\cite{hpsd}. \textsc{STAS} also depends on the base model: its configuration may require minor adaptation across models, and it cannot fully recover severely degraded generations. We discuss more in Appendix~\ref{app:limitations}.

\section{Conclusion}
In this paper, we investigate Massive Activations in video DiTs and uncover an emergent structural bias aligned with video-VAE encoding asymmetries: MA magnitudes peak at first-frame tokens and recur
at latent-frame boundaries, where the encoded content load is reduced. Bidirectional interventions further reveal that MAs act as implicit token-level rescalers, with larger magnitudes suppressing
self-attention and feed-forward residual writes. These findings suggest that video DiTs have learned to regulate residual computation according to the structural reliability of their latent tokens.
Building on this native behavior, we propose \textsc{STAS}, a training-free method that explicitly strengthens it by calibrating MAs at the identified structural positions during early denoising. Across diverse
backbones, \textsc{STAS} consistently improves video quality and temporal consistency without additional forward passes, while remaining compatible with existing training-free methods. Our results establish structured MAs as both an interpretable internal mechanism and a practical signal for improving video generation.

\bibliography{iclr2027_conference}
\bibliographystyle{iclr2027_conference}

\appendix
\clearpage

\noindent
\section*{Appendix}
\vspace{-8pt}
\section*{Steering Video Diffusion Transformers with Massive Activations}


\section{Massive Activation Analysis}
\subsection{Measurement Protocol}
\label{app:ma_measurement}
To identify the MA dimensions of different video DiT models, we compute the average absolute activations over 100 prompts using a streaming pass (running sums), at a fixed set of transformer blocks (and, when needed, a fixed subset of denoising steps). For each block/step, we first take the maximum absolute activation over all token positions for each feature dimension, and then take the mean of these per-dimension maxima as a reference scale. We then report MAs using a peak-to-mean rule: a dimension is marked as MA dimension only if its peak exceeds the mean reference by more than 50$\times$. For a new model, this identification requires only a one-pass activation profiling procedure, which means that activations are collected during ordinary forward evaluations and selected directly using the fixed criterion, without backpropagation, parameter search, or task-specific tuning. The profiling cost is therefore small and incurred only once per model.

\subsection{Statistics Across Models}
We summarize representative MA statistics at block 15 and timestep 5 across four models in Table \ref{tab:ma_weakma_t5}. For each model, we take the maximum activation magnitude over token positions for every feature dimension, and treat dimensions whose peaks exceed $\mu+3\sigma$ (computed over dimensions) as potential MA outliers. Among these candidates, dimensions with a peak-to-mean ratio larger than $50\times$ are marked as MA dimensions, while the remaining candidates are reported as weak-MA, indicating that they are elevated but less significant than the MA dims. As shown in Table \ref{tab:ma_weakma_t5}, different models have different MA/weak-MA dimensions. 

\begin{table}[h]
\centering
\scriptsize
\caption{\textbf{Model-specific massive activation dimensions across video DiTs.}}
\setlength{\tabcolsep}{4pt}
\begin{tabular}{lccccccc}
\toprule
Model & Block & $k$ & Type & Dim & Max Value & Peak-to-Mean & Peak-to-Median\\
\midrule
\rowcolor{orange!15}
Wan2.1--1.3B \cite{wan}  & 15 & 5 & MA      & \underline{\textbf{1188}} & \textbf{45.4400}  & \textbf{58.3}  & \textbf{59.6} \\
Wan2.1--1.3B \cite{wan}  & 15 & 5 & weak-MA &   \underline{71} & 4.8645   & 6.2   & 7.5  \\
\midrule
\rowcolor{orange!15}
CogVideoX--5B \cite{cogvideox}  & 15 & 5 & MA      & \underline{\textbf{1982}} & \textbf{412.5200} & \textbf{100.1} & \textbf{127.3} \\
\midrule
\rowcolor{orange!15}
Wan2.2--5B \cite{wan}    & 15 & 5 & MA      & \underline{\textbf{1938}} & \textbf{63.3419}  & \textbf{83.8}  & \textbf{97.1} \\
Wan2.2--5B \cite{wan}    & 15 & 5 & weak-MA & \underline{1389} & 5.0704   & 6.7   & 7.8  \\
Wan2.2--5B \cite{wan}    & 15 & 5 & weak-MA & \underline{2357} & 4.4530   & 5.9   & 6.8  \\
\midrule
\rowcolor{orange!15}
LTX2(base) \cite{ltx2}    & 15 & 5 & MA      & \underline{\textbf{3999}} & \textbf{321.0118}  & \textbf{110.2}  & \textbf{201.2} \\
LTX2(base) \cite{ltx2}    & 15 & 5 & weak-MA & \underline{1076} & 50.6361   & 17.4   & 31.7  \\
\bottomrule
\end{tabular}
\label{tab:ma_weakma_t5}
\end{table}

\section{STAS Implementation and Reproducibility Details}
\subsection{Algorithm}
\label{app:algorithm}
Algorithm~\ref{alg:stas} summarizes the complete inference procedure of \textsc{STAS}. Working with CFG, \textsc{STAS} calibrates the MA dimensions at selected structural tokens, DiT blocks, and early denoising steps using the max-referenced target. The operation is implemented as an in-place masked assignment over a small subset of hidden-state entries and requires neither additional denoiser evaluations nor auxiliary branches.

\begin{algorithm}[t]
  \caption{Structured Activation Steering (\textsc{STAS})}
  \label{alg:stas}
  \begin{algorithmic}[1]
  \Require Video DiT $D_\theta$, VAE decoder $\mathcal{D}$
  \Require Sampling schedule $\{t_k\}_{k=0}^{N-1}$ and CFG scale $\lambda$
  \Require MA dimensions $\mathcal{M}$ and intervention blocks $\mathcal{L}$
  \Require Steering strength $\alpha$, boundary ratio $p$, and step window $K$

  \State Sample $z_0\sim\mathcal{N}(0,I)$
  \State $\mathcal{S}\gets\mathcal{F}_0\cup\mathcal{B}(p)$

  \Function{Steer}{$H^m,k$}
      \If{$k<K$ \textbf{and} $m\in\mathcal{L}$}
          \For{each $d\in\mathcal{M}$}
              \State $r^m_{k,d}\gets
              \max_{j\in\mathcal{V}}\left|H^m_{k,j,d}\right|$
              \For{each $i\in\mathcal{S}$}
                  \State $H^m_{k,i,d}\gets
                  \alpha \cdot r^m_{k,d}\,
                  \cdot\operatorname{sign}\!\left(H^m_{k,i,d}\right)$
              \EndFor
          \EndFor
      \EndIf
      \State \Return $H^m$
  \EndFunction

  \For{$k=0,1,\ldots,N-1$}
      \State $v_k^{(+)}
      \gets D_\theta
      \bigl(z_k,t_k,c;\Call{Steer}{\cdot,k}\bigr)$
      \Comment{Conditional branch}

      \State $v_k^{(-)}
      \gets D_\theta
      \bigl(z_k,t_k,c_{\mathrm{neg}};\Call{Steer}{\cdot,k}\bigr)$
      \Comment{Unconditional branch}

      \State $v_k^{\mathrm{cfg}}
      \gets v_k^{(-)}
      +\lambda\!\left(v_k^{(+)}-v_k^{(-)}\right)$

      \State $z_{k+1}
      \gets \Call{Update}{z_k,v_k^{\mathrm{cfg}},t_k}$
  \EndFor

  \State \Return $\mathcal{D}(z_N)$
  \end{algorithmic}
  \end{algorithm}

\subsection{Model-Specific Configurations}
\label{app:model_configs}
We summarize the backbone configurations and the corresponding \textsc{STAS} settings in Table \ref{tab:model_cofig}. For each model, we report the number of transformer blocks, hidden size, the identified MA dimensions, and the layer index where \textsc{STAS} is applied. These configurations are used consistently across all experiments unless otherwise specified.
\begin{table}[th]
\centering
\small
\setlength{\tabcolsep}{6pt}
\caption{\textbf{Model configurations used for implementing \textsc{STAS} across backbones.}}
\begin{tabular}{lcccc}
\toprule
{Model} 
& {Blocks} 
& {Hidden size} 
& {MA dim} 
& {Applied layer} \\
\midrule
Wan2.1-1.3B    & 30 & 1536 & 1188          & 9  \\
Wan2.2-5B & 30 & 3072 & 1938         & 9 \\
CogVideoX-5B   & 42 & 3072 & 1982  & 8 \\
LTX2 (base)   & 48 & 4096 & 3999  & 7 \\
\bottomrule
\end{tabular}
\label{tab:model_cofig}
\end{table}

\begin{table}[!h]
\centering
\small
\setlength{\tabcolsep}{5.5pt}
\caption{\textbf{Hyperparameter setup across models.}}
\begin{tabular}{lcccc}
\toprule
{Model} 
& $\lambda$ 
& $\alpha$ 
& $p$ 
& $K$ \\
& (CFG scale) 
& (STAS scale) 
& (boundary \%) 
& (timestep cut-off) \\
\midrule
Wan2.1-1.3B    & 5.0 & 2.5 & 8 & 20  \\
Wan2.2-5B & 5.0 & 2.0 & 12 & 20 \\
CogVideoX-5B   & 6.0 & 1.2 & 8  & 20 \\
LTX2 (base)   & 3.0 & 1.2 & 8  & 30 \\
\bottomrule
\end{tabular}
\label{tab:hyper}
\end{table}

Since \textsc{STAS} is applied on top of classifier-free guidance (CFG)~\cite{cfg}. We use the default guidance scale $\lambda$ for each model. The \textsc{STAS} scaling factor $\alpha$, latent-boundary percentage $p$, and denoising timestep threshold $K$ are summarized in Table \ref{tab:hyper}.

\subsection{Experimental Protocol for Main Figures}
\label{app:experimental_protocal}
In this subsection, we detail the experimental settings and implementation specifics used to generate Figures \ref{fig:fig1}, \ref{fig:vae_border} and \ref{fig:rescale}.

\paragraph{Figure 1.} For all the 3D bar charts in Figure \ref{fig:fig1}, the z-axis values (activation magnitudes of hidden states) are collected from block 15 at an early denoising timestep ($k=5$, counting from the start of the denoising process). The corresponding generated video resolutions are $81\times480\times832$, $81\times704\times1280$, $49\times480\times720$, and $81\times512\times768$, respectively. 

\paragraph{Figure 2.} All VAE experiments in Figure~\ref{fig:vae_border} use the Wan2.1 video VAE
  on videos of resolution $T\times H_x\times W_x=81\times480\times832$, producing
  21 latent frames. We disable tiled encoding to avoid additional tile-boundary
  effects.

  For the temporal analysis in Figure~\ref{fig:vae_border_a} (the purple curve), we quantify the
  content carried by each latent frame through a decoder-ablation probe. Given
  video $V^{(q)}$ and its VAE latent
  $z^{(q)}=\mathcal{E}(V^{(q)})$, we ablate latent frame $f$ by replacing its
  spatially varying features with their per-channel spatial means:
  \begin{equation}
  \widetilde{z}^{(q,f)}_{c,\ell,h,w}
  =
  \begin{cases}
  \displaystyle
  \frac{1}{H_zW_z}
  \sum_{h',w'}z^{(q)}_{c,f,h',w'}, & \ell=f,\\[6pt]
  z^{(q)}_{c,\ell,h,w}, & \ell\neq f.
  \end{cases}
  \label{eq:latent_frame_ablation}
  \end{equation}
  Here, $f\in[0,20]$ denotes the ablated latent frame, whereas $\ell$ indexes an arbitrary
  latent frame. This operation preserves the channel-wise mean while removing
  the frame's spatial structure. We decode the original and ablated latents and
  measure the resulting change over the physical frames associated with $f$:
  \begin{equation}
  \mathcal{L}_f^{(q)}
  =
  \frac{1}{3H_xW_x}
  \sum_{\tau\in\mathcal{G}_f}
  \sum_{c,h,w}
  \left|
  \mathcal{D}(z^{(q)})_{c,\tau,h,w}
  -
  \mathcal{D}(\widetilde{z}^{(q,f)})_{c,\tau,h,w}
  \right|,
  \label{eq:latent_content_load}
  \end{equation}
  where $\mathcal{G}_0$ contains the first decoded physical frame, while
  $\mathcal{G}_f$ contains the four physical frames represented by each later
  latent frame. The purple curve plots, at each latent-frame index $f$, the median across all tested 
  videos of
  $100\,\mathcal{L}_f^{(q)}/\operatorname{median}_{r=1,\ldots,20}
  \mathcal{L}_r^{(q)}$.
  Hence, the value of approximately $26\%$ at $f=0$ means that the first latent
  frame has about one quarter of the total decoded influence of a typical
  subsequent latent frame. 

  For comparison in Figure \ref{fig:vae_border_a} (the orange curve), we collect the absolute activation of MA dimension 1188 from
  block 15 at sampling step $k=5$ over 100 prompt-conditioned generations.
  For each generation and latent-frame index $f$, we first average the absolute
  MA magnitude over the frame's interior tokens, excluding a two-token margin
  from every spatial boundary. The orange curve plots the mean across generations
  of this frame-wise magnitude divided by its median over all latent frames after
  the first and multiplied by $100$. Thus, a value of $119\%$ at $f=0$ means
  that the first latent frame has an average MA magnitude approximately $19\%$
  larger than a typical subsequent latent frame. 

  For the spatial analysis in Figure~\ref{fig:vae_border_b}, we construct eight
  control videos whose frames are spatially uniform. Their intensities span
  $[-0.9,0.9]$, with a small temporally varying but spatially constant
  perturbation. Because the inputs contain no spatial structure, systematic
  spatial variation in their latents reflects the VAE encoder rather than image
  content. We average the absolute latent values over $Q$ videos, $C_z$ channels, and all
  latent frames after the first, producing a $60\times104$ spatial map:
  \begin{equation}
  M^{\mathrm{VAE}}_{h,w}
  =
  \frac{1}{Q\,C_z\,(F_z-1)}
  \sum_{q=1}^{Q}
  \sum_{c=1}^{C_z}
  \sum_{f=1}^{F_z-1}
  \left|z^{(q)}_{c,f,h,w}\right|.
  \label{eq:vae_spatial_map}
  \end{equation}
  Excluding the first latent frame separates the spatial-padding effect from the
  distinct temporal-padding effect.

  The corresponding DiT map is computed from MA dimension 1188 at block 15 and
  sampling step $k=5$ over $Q_{\mathrm{MA}}=100$ generations. We reshape the flattened video
  tokens into a $21\times30\times52$ grid and average their absolute MA
  magnitudes over generations and latent frames $f\geq2$:
  \begin{equation}
  M^{\mathrm{MA}}_{h,w}
  =
  \frac{1}{Q_{\mathrm{MA}}(F_z-2)}
  \sum_{q=1}^{Q_{\mathrm{MA}}}
  \sum_{f=2}^{F_z-1}
  \left|
  H^{15,(q)}_{k,(f,h,w),1188}
  \right|.
  \label{eq:ma_spatial_map}
  \end{equation}
  The earliest two latent frames are excluded here to suppress the dominant temporal
  transient and isolate the recurring spatial pattern. After $2\times2$ spatial patchification, the $60\times104$ VAE latent grid becomes a $30\times52$ DiT token grid for each latent frame.

  For visualization, each map is independently normalized by the median of its
  interior positions, excluding the outermost spatial boundaries, and shown as
  the percentage deviation from this reference. Accordingly, $0\%$ denotes the
  interior median, while positive and negative values indicate larger and smaller
  magnitudes, respectively.

\paragraph{Figure 4.}
\label{app:rescale}
  For the Figure \ref{fig:rescale}, over 100 paired Wan2.1-1.3B generations, we steer MA dimension 1188 at the
  selected structural tokens in block 9 and measure the response in block 10 during
  the first ten denoising steps. We report post-normalization content and
  self-attention/FFN residual-write magnitudes, normalized by their paired native
  values. Dashed curves show the corresponding responses at non-target interior
  tokens.

\section{Additional Analysis of the VAE--MA Connection}
\label{app:additional_vae_ma}

\begin{figure}[t]
  \centering
  \includegraphics[width=0.8\linewidth]{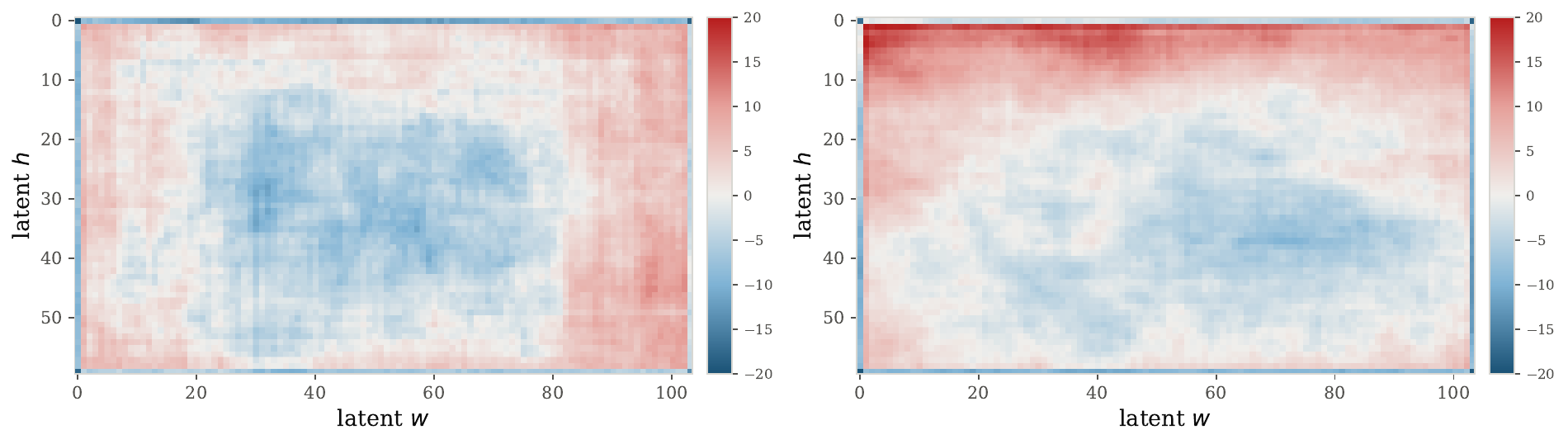}
  \caption{\textbf{VAE boundary responses on nonuniform inputs.}
  Natural videos are shown on the left and model generated videos on the right.}
  \label{fig:appendix_real_vs_uniform}
\end{figure}

\paragraph{VAE Boundary Responses on Natural and Generated Videos.} The spatial boundary response is not restricted to the uniform controls used in our main analysis. Figure~\ref{fig:appendix_real_vs_uniform} shows the average VAE latent magnitude for natural videos and generated videos (averaged over 100 videos each.). Both exhibit content-dependent, low-frequency spatial variation, yet retain a sharp one-cell attenuation at the outermost latent boundary. Thus, the boundary signature introduced by the VAE remains visible under realistic content and in the model's own outputs. Its alignment with elevated MAs at the same positions, further supporting the VAE--MA connection.

\begin{figure}[t]
    \centering
    \begin{subfigure}[b]{0.31\linewidth}
        \centering
        \includegraphics[width=\linewidth]{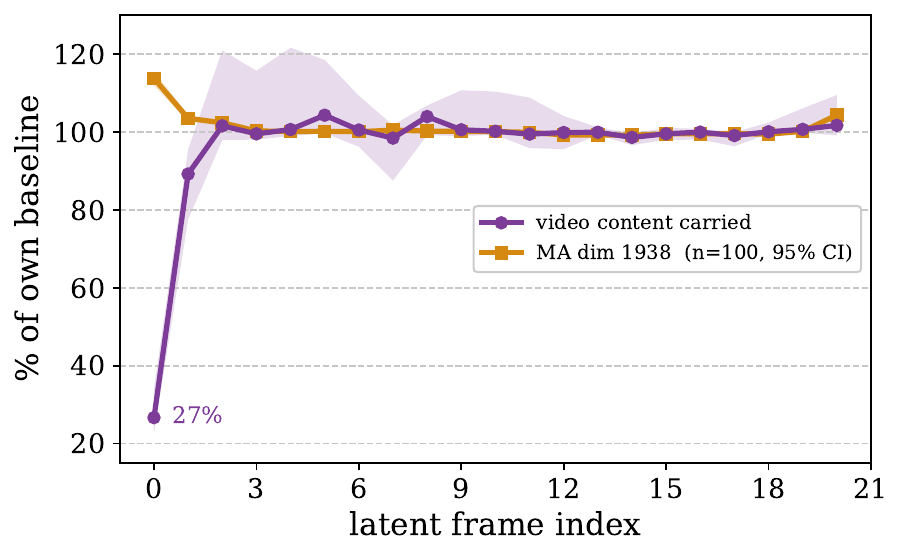}
        \caption{Per-frame content load vs.\ MA}
    \end{subfigure}
    \hfill
    \begin{subfigure}[b]{0.675\linewidth}
        \centering
        \includegraphics[width=\linewidth]{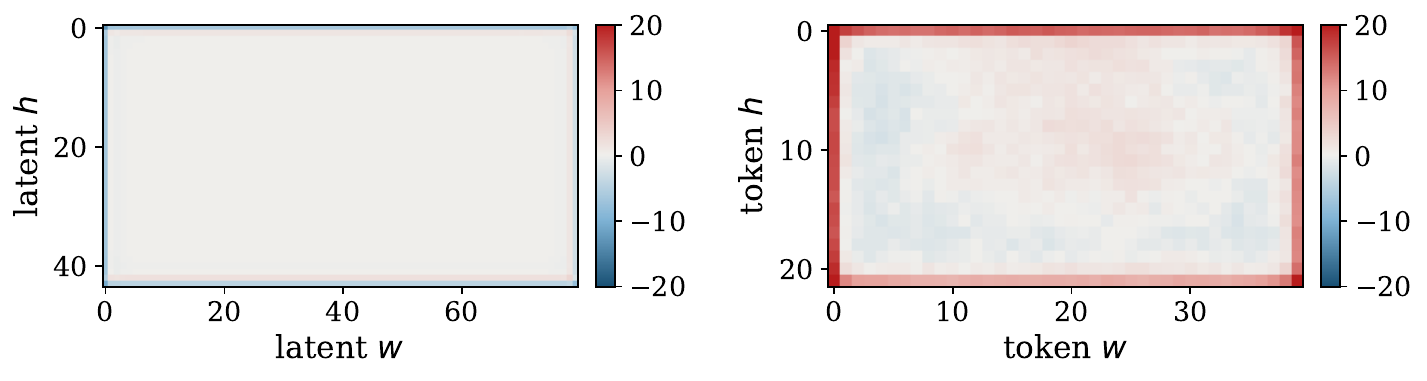}
        \caption{Latent boundaries of the VAE (left) and the DiT MA map (right)}
    \end{subfigure}
  \caption{\textbf{The MA hierarchy aligns with video-VAE encoding asymmetries on Wan2.2 VAE.}}
    \label{fig:wan22_vae_border}
\end{figure}

\begin{figure}[!t]
    \centering
    \begin{subfigure}[b]{0.31\linewidth}
        \centering
        \includegraphics[width=\linewidth]{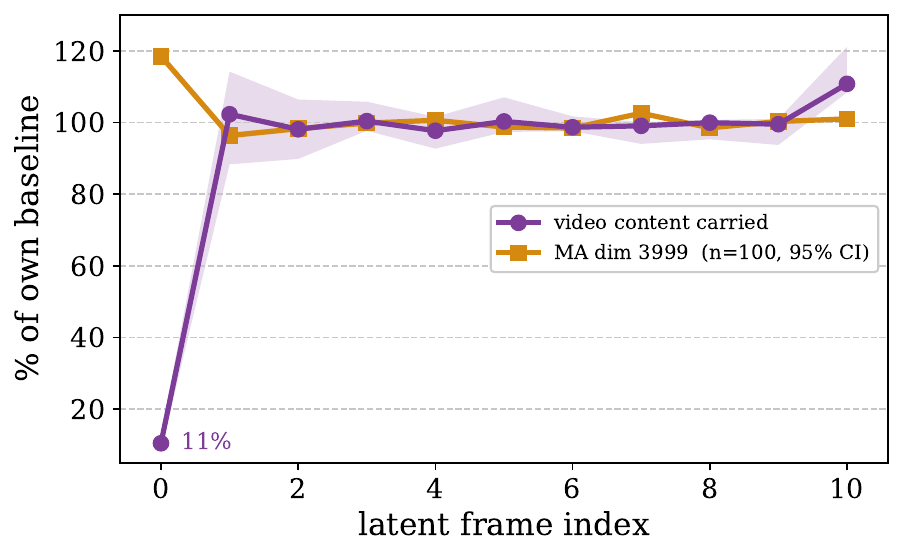}
        \caption{Per-frame content load vs.\ MA}
    \end{subfigure}
    \hfill
    \begin{subfigure}[b]{0.645\linewidth}
        \centering
        \includegraphics[width=\linewidth]{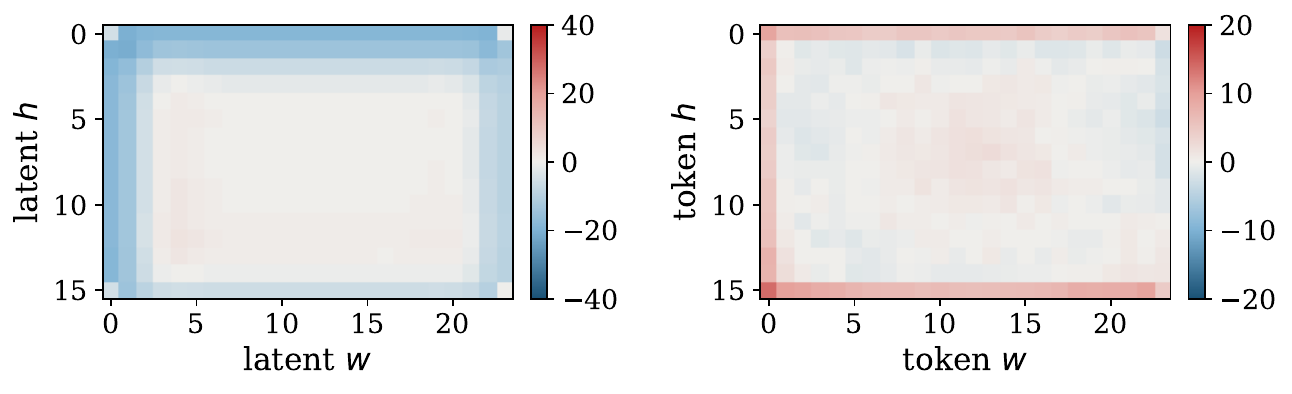}
        \caption{Latent boundaries of the VAE (left) and the DiT MA map (right)}
    \end{subfigure}
  \caption{\textbf{The MA hierarchy aligns with video-VAE encoding asymmetries on LTX-2 VAE.}}
    \label{fig:ltx2_vae_border}
\end{figure}

\paragraph{Generalization across different VAEs.} We further extend the VAE--MA correspondence experiments on Wan2.2 and LTX2 video VAEs, which use different spatial and temporal compresion ratios shown in Figures \ref{fig:wan22_vae_border} and \ref{fig:ltx2_vae_border}. Temporally, the first latent frame carries substantially less video content than later frames, while exhibiting a consistently higher MA magnitude. Spatially, both models exhibit attenuated VAE latent magnitude at boundary positions alongside elevated MAs at the corresponding tokens. These results show that the inverse relationship between VAE content load and MA magnitude generalizes across different VAEs.

\section{Analysis of the Role of RoPE in Localizing MA Patterns}
\label{app:rope}
Asymmetric spatial padding and causal temporal processing in the video VAE
produce a systematic information asymmetry between boundary and interior latent tokens. Because the same VAE transformation is applied throughout training, this pattern is expected to persist in the latent training distribution. We hypothesize a three-stage account: the VAE creates the asymmetry, the DiT learns its recurring structure, and RoPE supplies the coordinates that localize the learned response in the flattened token sequence. We test the localization stage by modifying only the RoPE lookup and no latent token is explicitly moved or overwritten.

Specifically, we use Wan2.1-T2V-1.3B and follow the MA identification protocol described in Appendix~\ref{app:ma_measurement}. All conditions use the same 100 prompts, random seeds, model weights, and sampling settings. We record activations at block 15 and timestep 5. An 81-frame video corresponds to 21 latent frames of $30\times52$ tokens, leading to $1{,}560$ consecutive positions in the flattened token sequence.


\begin{figure}[t]
  \centering

  \begin{subfigure}[t]{0.24\linewidth}
      \centering
      \includegraphics[width=\linewidth]{figures/wan2.1_b15_t5.png}
      \caption{Native RoPE}
      \label{fig:app_rope_fig1a}
  \end{subfigure}\hfill
  \begin{subfigure}[t]{0.24\linewidth}
      \centering
      \includegraphics[width=\linewidth]{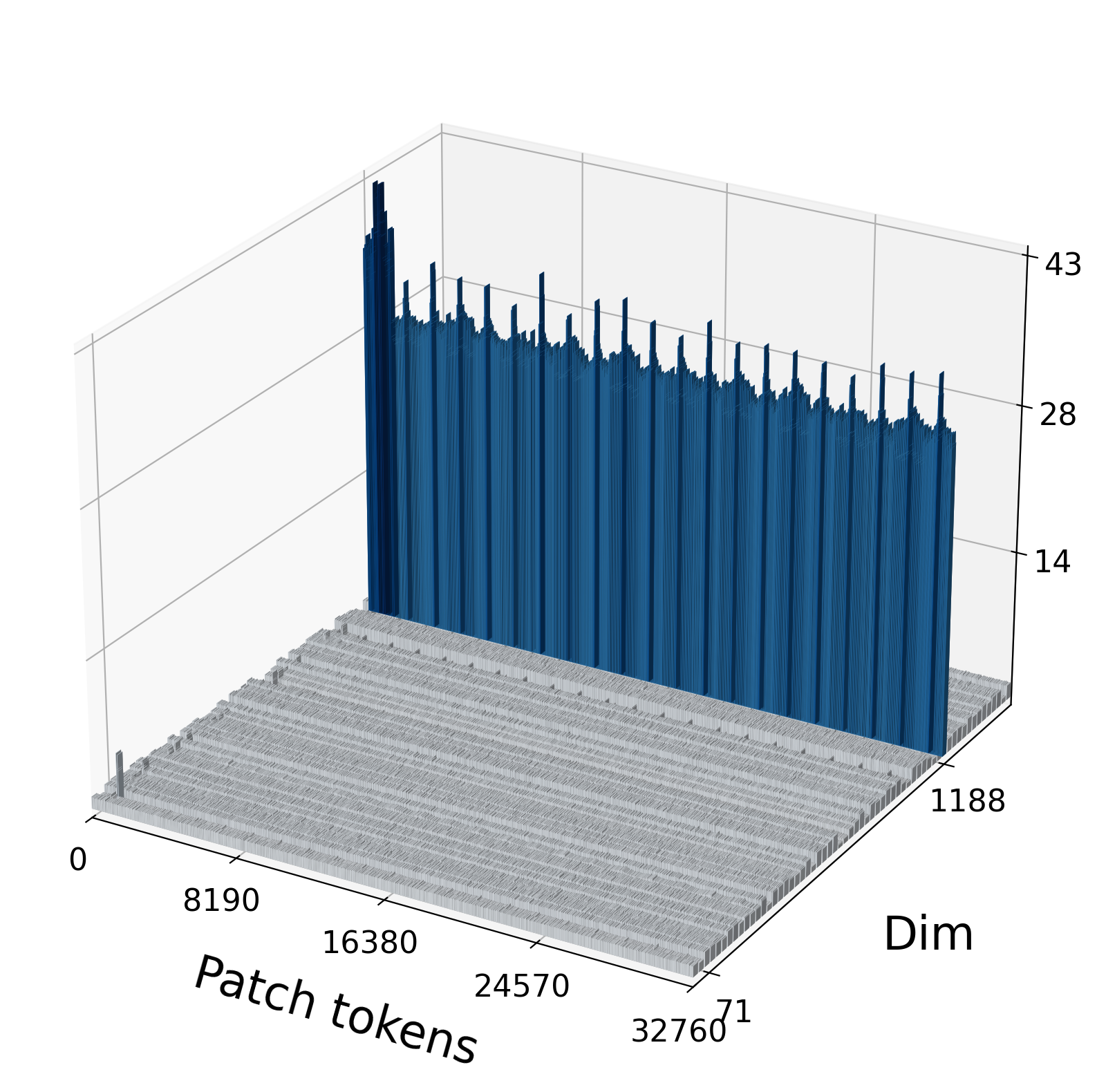}
      \caption{Height roll (+15)}
      \label{fig:app_rope_fig1b}
  \end{subfigure}\hfill
  \begin{subfigure}[t]{0.24\linewidth}
      \centering
      \includegraphics[width=\linewidth]{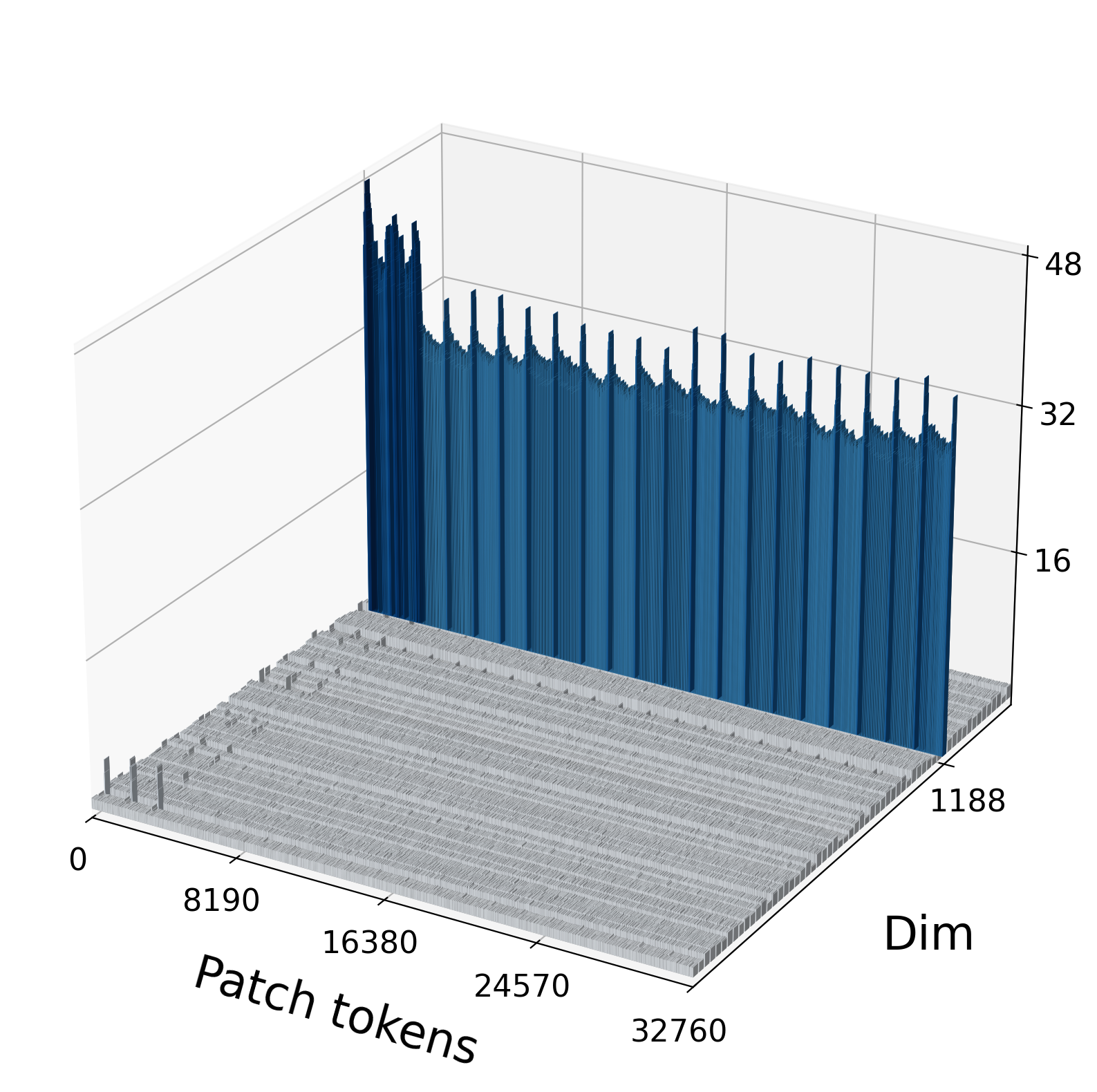}
      \caption{Temporal share, $G{=}2$}
      \label{fig:app_rope_fig1c}
  \end{subfigure}\hfill
  \begin{subfigure}[t]{0.24\linewidth}
      \centering
      \includegraphics[width=\linewidth]{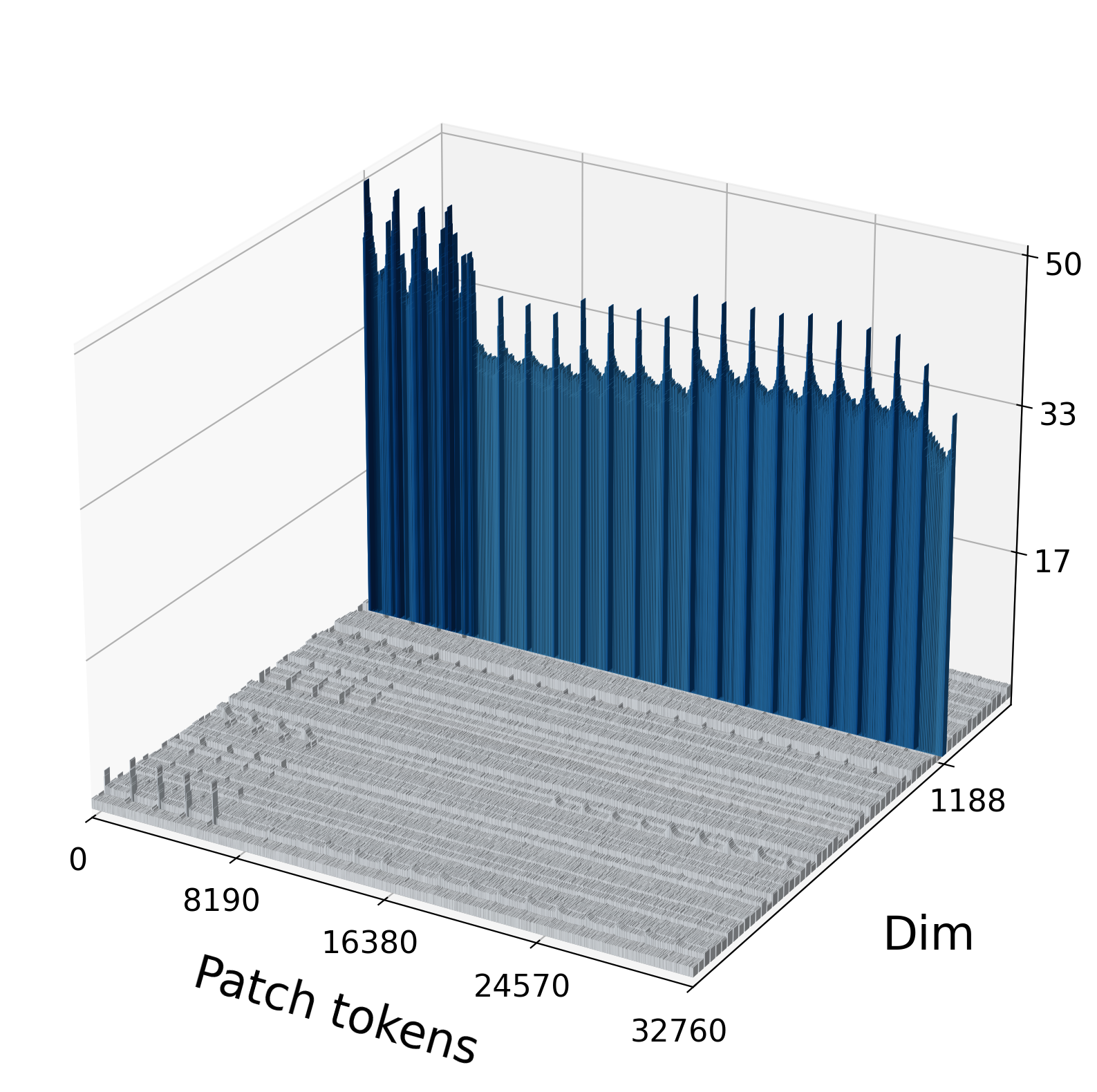}
      \caption{Temporal share, $G{=}4$}
      \label{fig:app_rope_fig1d}
  \end{subfigure}

  \medskip

  \begin{subfigure}[t]{0.44\linewidth}
      \centering
      \includegraphics[width=\linewidth]{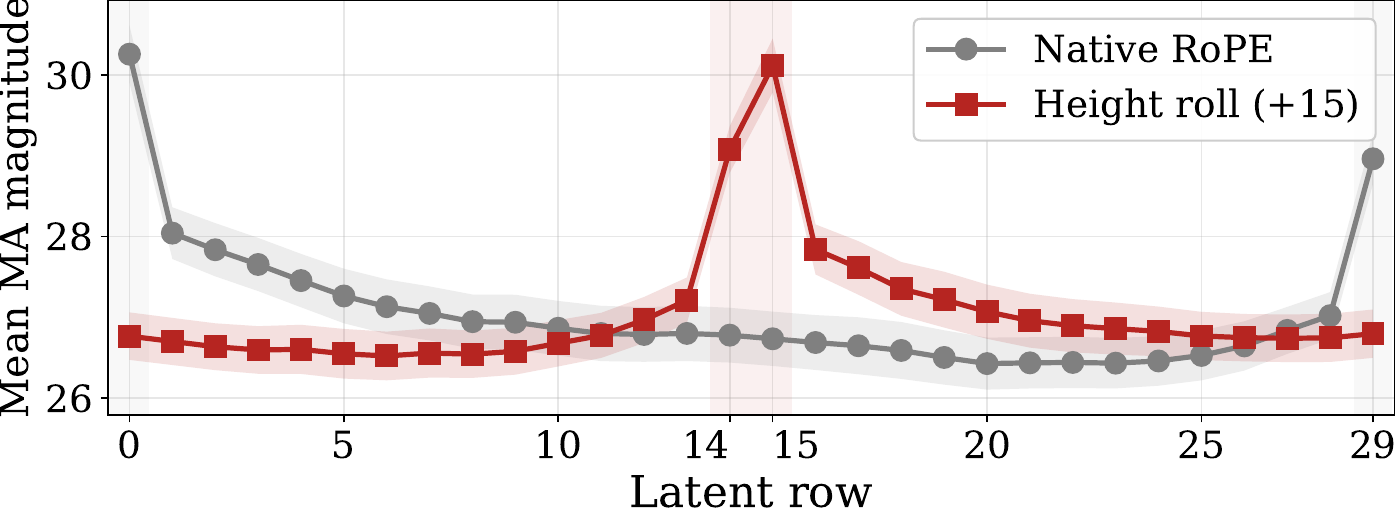}
      \caption{Spatial MA profiles}
      \label{fig:app_rope_fig1e}
  \end{subfigure}\hspace{0.02\linewidth}
  \begin{subfigure}[t]{0.44\linewidth}
      \centering
      \includegraphics[width=\linewidth]{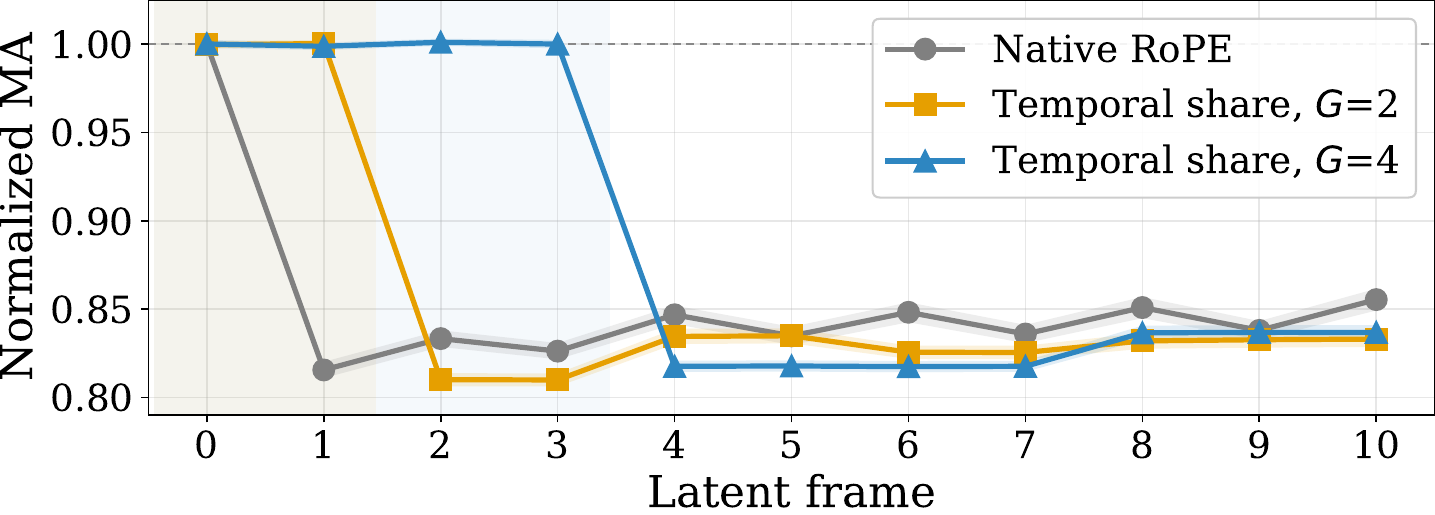}
      \caption{Temporal MA profiles}
      \label{fig:app_rope_fig1f}
  \end{subfigure}

  \caption{
  \textbf{MA peaks follow the RoPE coordinates.}
  (a) Native RoPE.
  (b) Rolling the height RoPE indices by ${+}15$ moves the boundary
  coordinates to the center rows, and the MA peaks move with them.
  (c--d) Sharing each temporal RoPE coordinate across groups of
  $G{=}2$ or $G{=}4$ consecutive latent frames makes the first $G$
  frames share the first-frame coordinate and exhibit a common elevated
  MA plateau.
  (e) The aggregated spatial profiles show the MA peak moving from the
  boundary rows to the center rows under the height roll.
  (f) The normalized temporal profiles show elevated plateaus spanning
  one, two, and four latent frames under native RoPE, $G{=}2$, and
  $G{=}4$, respectively.
  }
  \label{fig:appendix_rope_ma}
  \end{figure}

\paragraph{Spatial RoPE roll.}
We first shift every height coordinate by half the grid in RoPE:
\begin{equation}
    \pi_h(r)=(r+15)\bmod 30 .
\end{equation}
The boundary rows $0$ and $29$ thus receive central indices and the central rows $14$ and $15$ receive boundary indices, with the native order
$0,1,\dots,29$ becoming $15,16,\dots,29,0,1,\dots,14$. The shift is a bijection on the index set, and the token order and all other
coordinates are unchanged.

With 52 tokens per row, shifting by 15 rows predicts a displacement of
$15\times52=780$ tokens, exactly half the $1{,}560$-token frame period. This is what we observe in Fig.~\ref{fig:app_rope_fig1b}: in every latent frame, including the first, the peaks move off the physical boundary rows and toward the frame center. The aggregated row profiles in Fig.~\ref{fig:app_rope_fig1e} further make this relocation explicit.

\paragraph{Temporal RoPE grouping.}
  Native indexing assigns latent frame $i$ the temporal coordinate
  $\pi_t(i)=i$. We next use
  \begin{equation}
      \pi_t^{(G)}(i)
      =
      \left\lfloor \frac{i}{G} \right\rfloor,
      \qquad G\in\{2,4\},
  \end{equation}
so that $G$ consecutive latent frames share one temporal coordinate, while the height and width coordinates remain unchanged. For $G=2$ the native sequence $0,1,2,\dots,20$ becomes $0,0,1,1,\dots,9,9,10$, so the tokens of the first two frames all carry the temporal coordinate of the first frame. The first latent frame MA elevation should then cover both.

  Figures~\ref{fig:app_rope_fig1c} and~\ref{fig:app_rope_fig1d} show this behavior, in which the elevated region expands to two latent frames for $G=2$ and four latent frames for $G=4$. Consistently, the averaged profiles in Fig.~\ref{fig:app_rope_fig1f} contain elevated plateaus of 1, 2, and 4 latent frames under native indexing, $G=2$, and $G=4$, respectively. Although repeated temporal indices constitute an out-of-distribution coordinate assignment, the systematic widening follows the prediction of a coordinate-driven mechanism.

\paragraph{Interpretation.}
The two interventions probe complementary axes of the factorized RoPE.
Spatial half-period shift changes its phase within each frame, whereas
temporal grouping changes the extent of the MA pattern across latent frames.
Together, these results establish the assigned RoPE coordinates as the
localization mechanism for MA. RoPE transports the VAE-induced asymmetry to
the corresponding token positions, and the VAE generates the underlying
asymmetry.

\section{Additional Analysis of MA-Induced Residual Rescaling}
  \label{app:rescaler_math}

We give a normalization-based analysis of why MA magnitude rescales residual writing. Both LayerNorm and RMSNorm fix the overall scale of each token, so a single unusually large hidden dimension inflates the normalization denominator and shrinks the relative contribution of all remaining dimensions. We derive this effect for the normalization layers used by the studied backbones, omitting the stabilization constant $\epsilon$, which is negligible at the observed magnitudes.

Fix a block $m$, denoising step $k$, and token $i$, and abbreviate
$h_d = H^{m}_{k,i,d}$, where $C$ is the number of channels. Let
$d^* \in \mathcal{M}$ denote the dominant MA dimension, with signed
activation $a = h_{d^*}$ and magnitude $A = |a|$.

\paragraph{LayerNorm.}
  Wan2.1 and CogVideoX use LayerNorm before self-attention and feed-forward branch, which subtracts the token mean
  \(\mu\) from every dimension and divides by the token standard deviation
  \(\sqrt{v}\). To see how the MA affects this normalization, let \(\bar c\)
  be the mean of the remaining dimensions and let
  \(E_{\mathrm{LN}}=\sum_{d'\neq d^*}(h_{d'}-\bar c)^2\) denote their centered
  energy. Separating the MA dimension gives
  \begin{equation}
    v
    =
    \frac{1}{C}\sum_{d'=1}^{C}(h_{d'}-\mu)^2
    =
    \underbrace{\frac{E_{\mathrm{LN}}}{C}}_{\text{remaining dimensions}}
    +
    \underbrace{\frac{C-1}{C^2}(a-\bar c)^2}_{\text{MA contribution}}.
    \label{eq:ln_var}
  \end{equation}
  The first term is the variance already present among the remaining
  dimensions, while the second is the additional variance introduced by the
  MA. As \(\lvert a-\bar c\rvert\) grows, every dimension is divided by a
  larger standard deviation. Mean subtraction also shifts all non-MA
  dimensions by the same amount, but this shift cancels in their pairwise
  differences. Their contrast is therefore attenuated by the enlarged
  denominator. Since \(A\gg|\bar c|\) and \(C\) is large, this attenuation is
  approximately \((1+A^2/E_{\mathrm{LN}})^{-1/2}\), so that erasing the MA releases
  the remaining dimensions, whereas amplifying it suppresses them.

  \paragraph{RMSNorm.}
  LTX-2 divides each coordinate by the root-mean-square magnitude of the token.
  Writing \(E_{\mathrm{RMS}}=\sum_{d'\neq d^*}h_{d'}^2\) for the energy of the
  remaining dimensions, every \(d\neq d^*\) is mapped to
  \begin{equation}
    \operatorname{RMSNorm}(h)_d
    =
    \frac{h_d}{\sqrt{(E_{\mathrm{RMS}}+A^2)/C}}
    =
    \rho(A)\frac{h_d}{\sqrt{E_{\mathrm{RMS}}/C}},
    \qquad
    \rho(A)=\frac{1}{\sqrt{1+A^2/E_{\mathrm{RMS}}}}.
    \label{eq:rho_rms}
  \end{equation}
  RMSNorm therefore yields the same attenuation form directly. The factor
  \(\rho(A)\) equals one when zeroing out MA and decreases monotonically as
  \(A\) grows.

  LayerNorm and RMSNorm thus share the same mechanism: a dominant MA increases
  the token's normalization denominator, leaving less normalized scale for its
  remaining dimensions. LayerNorm measures their centered energy, whereas
  RMSNorm uses their uncentered energy. When several MA dimensions are active,
  their joint contribution enters the denominator and produces the same
  effect.


\paragraph{Implications for residual writing.}
As normalization is applied to each token independently and a token is divided by its own RMS magnitude, a larger MA inflates that token's denominator and leaves less normalized scale for its remaining dimensions. The attenuation therefore depends on the MA magnitude relative to the energy \(E_{\mathrm{RMS}}\) of those dimensions. Scaling the MA dimension alone does not change \(E_{\mathrm{RMS}}\), so \(\rho\) decreases strictly monotonically in
\(A\) at that token. The normalized token then feeds both the attention-value and feed-forward projections. Amplifying a token's MA therefore weakens its residual writes, whereas erasing it strengthens them.

Interestingly, although the feed-forward branch processes each token independently, self-attention mixes tokens and can propagate the intervention to non-target positions. After projection, queries and keys are normalized
again by QK normalization before the attention scores are computed. This equalizes their vector lengths but does not restore their original coordinate values. Consequently, the query--key dot products and attention weights can still change. Figure~\ref{fig:rescale} supports this explanation. At selected tokens, MA erasure strengthens both residual branches, while amplification progressively suppresses them. At non-target tokens, the main response occurs in the self-attention write, with feed-forward writes remaining close to their native values. Affine and AdaLN parameters are shared across video tokens at a given block and denoising step, so they can change absolute magnitudes but do not eliminate this token-dependent attenuation.

\section{Additional Analysis of STAS Design Choices}
\label{app:coverage}

\begin{wrapfigure}{r}{0.38\linewidth}
\vspace{-12pt}
\centering
\includegraphics[width=\linewidth]{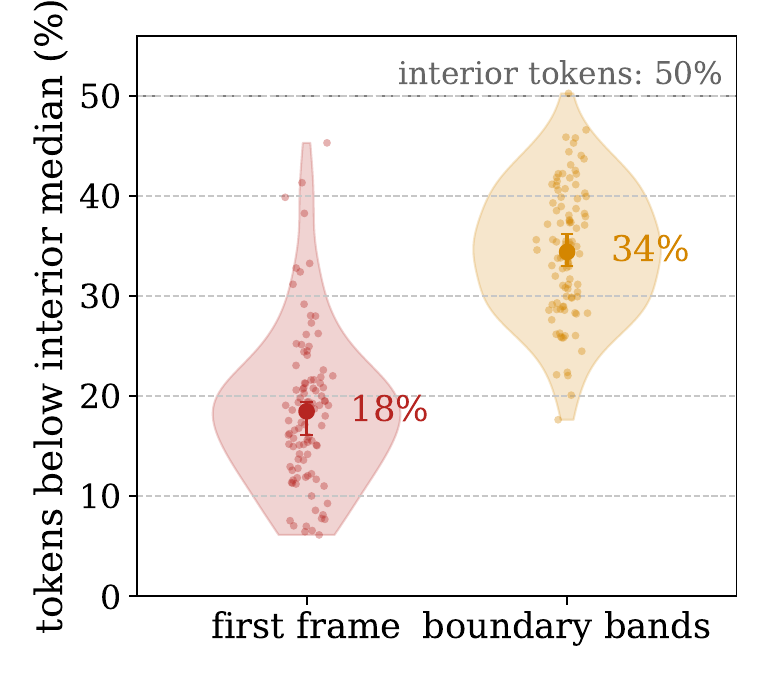}
\caption{\textbf{Structural MAs are not uniformly elevated.}}
\label{fig:ma_coverage}
\vspace{-14pt}
\end{wrapfigure}

Despite structural token positions exhibiting larger MA magnitudes on average, as discussed in Sec.~\ref{sec:observations}, this elevation does not hold for every token: an individual structural token may still have an MA magnitude no greater than that of a typical interior token. To quantify this overlap, we measure the fraction of structural tokens whose native MA falls below the interior-token median, across 100 generated videos. This fraction would be \(50\%\) if structural and interior tokens followed the same distribution, whereas \(0\%\) would mean that every structural token has a larger MA than the interior median. Figure~\ref{fig:ma_coverage} shows values of \(18\%\) for first-frame tokens and \(34\%\) for boundary-band tokens, indicating that structural MAs are generally stronger but remain heterogeneous.

Recall that our steering target in Eq.~\ref{eq:stas_target} determines the magnitude of each selected activation using the maximum response of the corresponding dimension while preserving its original
polarity. Since this target is independent of the token's native magnitude, it brings both weak and strong structural responses to a consistent adaptive scale. Simple multiplicative scaling instead
preserves their relative differences, leaving weak structural MAs comparatively weak. This explains why max-referenced calibration provides a better performance than scaling as shown in Figure \ref{fig:calibration_comparison} and Table \ref{tab:ablation}.

\section{Analysis of STAS Through the Lens of Guidance}
\label{app:guidance_view}
We analyze STAS through the lens of guidance by comparing it with
  CFG~\cite{cfg}, Autoguidance~\cite{karras2024guiding}, and
  DG~\cite{massive}. Each method can be written as an original quantity plus a
  correction. Specifically, CFG, Autoguidance, and DG correct the denoiser prediction, while
  \textsc{STAS} corrects internal hidden activations. To make this comparison explicit,
  let \(D=D_\theta(z_t,t,c)\) denote the original conditional denoiser
  prediction and define
  \[
  D_{\hat c}=D_\theta(z_t,t,\hat c),\qquad
  D_{\theta^*}=D_{\theta^*}(z_t,t,c),\qquad
  D_{\hat z}=D_\theta(\hat z_t,t,c).
  \]
  Here, \(\hat c\) denotes an unconditional or negative condition,
  \(\theta^*\) denotes the parameters of an under-capacity model, and
  \(\hat z_t\) denotes the MA-disrupted state used to construct the DG
  reference prediction.

  \textsc{STAS} does not combine two denoiser predictions. Instead, it directly replaces
  selected hidden activations with the calibrated target \(g\). To express this
  replacement as an additive update, let \(\chi^m_{k,i,d}\) indicate whether an
  activation is selected: it equals \(1\) when \(k<K\),
  \(m\in\mathcal L\), \(i\in\mathcal S\), and \(d\in\mathcal M\), and
  equals \(0\) otherwise. Equation~\ref{eq:stas} can then be rewritten as
  \begin{equation}
  \widehat H^m_{k,i,d}
  =
  H^m_{k,i,d}
  +
  \chi^m_{k,i,d}
  \left(g^m_{k,i,d}-H^m_{k,i,d}\right).
  \label{eq:stas_guidance}
  \end{equation}
  For a selected activation, \(\chi^m_{k,i,d}=1\), so the added term changes
  its value exactly from \(H^m_{k,i,d}\) to \(g^m_{k,i,d}\). For every other
  activation, \(\chi^m_{k,i,d}=0\), so no correction is applied.

  \begin{table*}[t]
  \centering
  \begingroup

  \caption{\textbf{Unified formulations of output guidance and STAS activation
  steering.}}
  \label{tab:guidance_comparison}

  \vspace{1pt}
  \scriptsize
  \setlength{\tabcolsep}{6pt}
  \renewcommand{\arraystretch}{1.08}

  \begin{tabular}{l c c l}
  \toprule
  Type
  & Formulation
  & Applied to
  & Reference / Target \\
  \midrule

  CFG
  & \(\widehat D=D+w\!\left(D-D_{\hat c}\right)\)
  & Denoiser output
  & Unconditional \(D_{\hat c}\) \\

  Autoguidance
  & \(\widehat D=D+w\!\left(D-D_{\theta^*}\right)\)
  & Denoiser output
  & Under-capacity \(D_{\theta^*}\) \\

  DG
  & \(\widehat D=D+w\!\left(D-D_{\hat z}\right)\)
  & Denoiser output
  & MA-disrupted \(D_{\hat z}\) \\

  \midrule

  \textbf{STAS}
  & \(\widehat H=H+\chi\!\left(g-H\right)\)
  & Hidden activation
  & MA-related adaptive target \(g\) \\

  \bottomrule
  \end{tabular}

  \endgroup
  \end{table*}

  Table~\ref{tab:guidance_comparison} highlights both the shared formulation and
  the key distinction. CFG, Autoguidance, and DG construct an auxiliary
  reference prediction and extrapolate away from it in denoiser-output space. CFG obtains its reference by removing the text condition, Autoguidance by replacing the denoiser with an under-capacity model, and DG by disrupting MAs in an auxiliary branch. Their guidance weight \(w\) controls the correction magnitude along the difference between the original and reference predictions.

  \textsc{STAS} instead operates directly in activation space. Rather than constructing an auxiliary prediction to extrapolate from, it rescales selected MA activations toward the max-referenced target \(g\), whose magnitude is controlled by \(\alpha\) in Equation~\ref{eq:stas_target}. MAs have previously been used as something to disrupt: DG perturbs them to obtain a degraded reference and then corrects the denoiser output. \textsc{STAS} instead treats them as a quantity to calibrate, adjusting them in place so that the corrected activations propagate through the remaining blocks. Consequently, unlike Autoguidance and DG, \textsc{STAS} requires neither an auxiliary model nor an additional degraded forward pass. It applies a single element-wise operation at a few structural token positions, adding no parameters and negligible computational overhead.

\section{Additional Results}
\label{app:additional_results}

\subsection{Additional Quantitative Results}
\label{app:more_quantitative}
Table~\ref{tab:vbench_full} presents the complete VBench breakdown over all 16 evaluation dimensions, together with the aggregated quality, semantic, and final scores across the four backbone models, showing consistent improvements across most dimensions.

\begin{table*}[t]
\centering
\caption{\textbf{Quantitative comparison on VBench.}}
\label{tab:vbench_full}
\footnotesize
\setlength{\tabcolsep}{6pt}
\begin{tabular}{
  l
  c >{\columncolor{orange!15}}c
  c >{\columncolor{orange!15}}c
  c >{\columncolor{orange!15}}c
  c >{\columncolor{orange!15}}c
  }
\toprule
\multirow{2}{*}{Dimension} 
& \multicolumn{2}{c}{Wan2.1-1.3B} 
& \multicolumn{2}{c}{CogVideoX-5B}
& \multicolumn{2}{c}{Wan2.2-5B} 
& \multicolumn{2}{c}{LTX-2 (base)} \\
\cmidrule(lr){2-3}
\cmidrule(lr){4-5}
\cmidrule(lr){6-7}
\cmidrule(lr){8-9}
& Vanilla & \textbf{+Ours} 
& Vanilla & \textbf{+Ours} 
& Vanilla & \textbf{+Ours} 
& Vanilla & \textbf{+Ours}  \\
\midrule
Subject Consistency   & 94.63 & \textbf{95.00} & 93.40 & \textbf{93.80} & 95.13 & \textbf{95.37} & 92.03 & \textbf{92.11}\\
BG. Consistency& 95.81 & \textbf{95.93} & 95.29 & \textbf{95.47} & 96.63 & \textbf{96.70} & 94.73 & \textbf{94.77}\\
Temporal Flickering   & 97.93 & \textbf{98.09} & 95.97 & \textbf{96.21} & 97.81 & \textbf{97.87} & \textbf{97.80} & \textbf{97.80}\\
Motion Smoothness     & 98.78 & \textbf{98.85} & 98.18 & \textbf{98.28} & \textbf{98.94} & \textbf{98.94} & \textbf{98.90} & \textbf{98.90}\\
Dynamic Degree        & \textbf{45.15} & 43.64 & \textbf{48.71} & 45.85 & \textbf{39.32} & 38.24 & \textbf{51.74} & 51.63\\
Aesthetic Quality     & 61.91 & \textbf{62.03} & 59.98 & \textbf{60.31} & 61.67 & \textbf{61.72} & \textbf{60.34} & \textbf{60.34}\\
Imaging Quality       & 68.14 & \textbf{68.95} & 64.62 & \textbf{65.12} & 69.02 & \textbf{69.39} & \textbf{66.36} & 66.35\\
Object Class          & 90.38 & \textbf{91.64} & 86.88 & \textbf{88.07} & \textbf{92.34} & 92.23 & \textbf{88.97} & 88.94\\
Multiple Objects      & 77.47 & \textbf{78.61} & 67.47 & \textbf{68.78} & 78.60 & \textbf{82.86} &64.59 & \textbf{66.14}\\
Human Action          & 96.40 & \textbf{97.20} & 98.60 & \textbf{98.80} & \textbf{98.00} & 97.80 & 98.20 & \textbf{98.40}\\
Color                 & \textbf{86.93} & 86.72 & 82.04 & \textbf{84.12} & 87.18 & \textbf{87.54} & 75.91 &\textbf{77.47}\\
Spatial Relationship  & 73.63 & \textbf{78.38} & 64.17 & \textbf{64.73} & 86.68 & \textbf{87.31} & \textbf{77.23} & 77.07\\
Scene                 & 54.87 & \textbf{55.41} & \textbf{54.62} & 54.41 & 54.18 & \textbf{55.47} & 54.74 & \textbf{55.71}\\
Appearance Style      & 22.90 & \textbf{22.93} & 24.83 & \textbf{24.92} & \textbf{22.84} & 22.77 & 22.67 & \textbf{22.75}\\
Temporal Style        & 25.97 & \textbf{26.02} & 25.49 & \textbf{25.59} & \textbf{25.81} & 25.80 &\textbf{24.62} & 24.59\\
Overall Consistency   & \textbf{27.03} & 27.02 & 27.59 & \textbf{27.65} & \textbf{27.50} & 27.44 & \textbf{26.57} & 26.54\\
\midrule
Quality Score & 81.81 & \textbf{82.03} & 79.78 & \textbf{79.95} & 81.75 & \textbf{81.82} & 81.11 &\textbf{81.13}\\
Semantic Score  & 79.70 & \textbf{80.66} & 77.59 & \textbf{78.24} & 81.68 & \textbf{82.35} & 76.82 & \textbf{77.32}\\
Final Score    & 81.39 & \textbf{81.76} & 79.34 & \textbf{79.61} & 81.74 & \textbf{81.93} &80.26 & \textbf{80.37}\\
\bottomrule
\end{tabular}
\end{table*}

\subsection{Additional Ablations}
\label{app:ablation}
We here provide additional ablation studies to further analyze more design choices in STAS: the boundary-percentage $p$, the scaling factor $\alpha$ and the applying layer depth $m$.

\begin{figure}[t]
    \centering
    \begin{subfigure}[t]{0.36\linewidth}
        \centering
        \includegraphics[width=\linewidth]{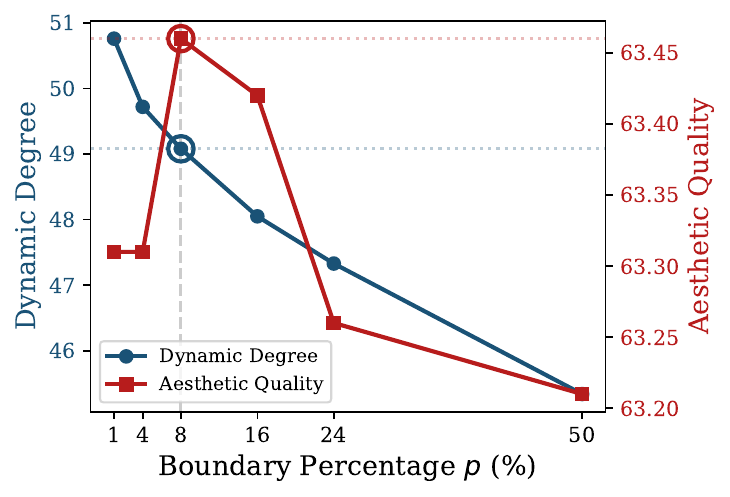}
        \caption{Effect of boundary percentages.}
        \label{fig:p}
    \end{subfigure}
    \begin{subfigure}[t]{0.36\linewidth}
        \centering
        \includegraphics[width=\linewidth]{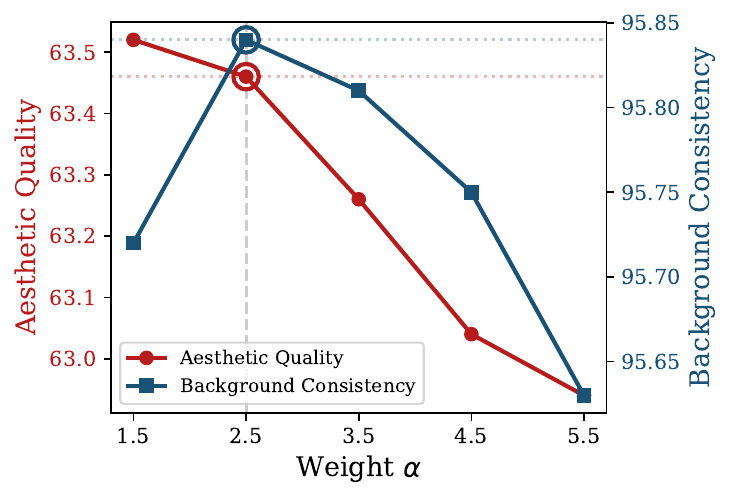}
        \caption{Effect of STAS weights.}
        \label{fig:alpha}
    \end{subfigure}
    \begin{subfigure}[t]{0.245\linewidth}
        \centering
        \includegraphics[width=\linewidth]{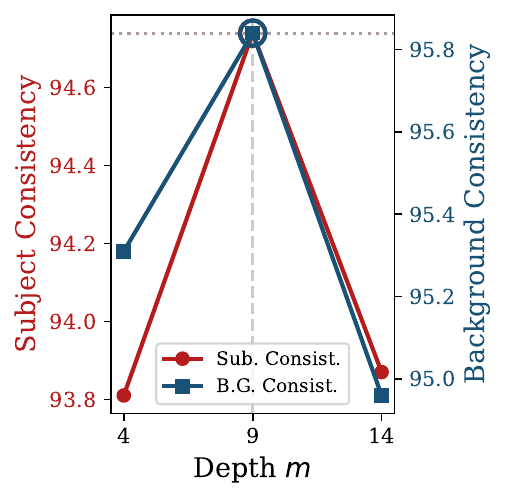}
        \caption{Effect of depths.}
        \label{fig:m}
    \end{subfigure}

    \caption{\textbf{More ablations.} Impact on boundary percentage $p$, STAS weight $\alpha$ and applied layer depth $m$. Moderate values achieve the best trade-off across metrics.}
    
    \label{fig:ablation_p_alpha}
\end{figure}

\paragraph{Effect of boundary percentage $p$.}
We present the quantitative results across different boundary percentage $p$ (head $p$\% + tail $p$\%) of latent chunks. As shown in Figure \ref{fig:p}, increasing $p$ from very small values initially improves aesthetic quality, reaching a peak at moderate percentages (around $8\%$), while the dynamic degree gradually decreases as $p$ becomes larger. Note that overly large $p$, which involves more interior tokens for calibration, would lead to a degradation for both metrics. For instance, when $p=50$, which all the visual tokens are selected for amplification, the performance is poor.
These results suggest that a moderate boundary coverage achieves the best trade-off.

\paragraph{Effect of scaling weight $\alpha$.}
We next show the effect of varying scaling weight $\alpha$, which controls the magnitude of amplification. 
As illustrated in Figure \ref{fig:alpha}, both metrics improve when increasing $\alpha$ from small values, peaking at a moderate weight (around $\alpha=2.5$). 
However, further increasing $\alpha$ leads to a consistent decline in both metrics.

\paragraph{Effect of applying layer depth $m$.}
We further ablate the transformer layer depth $m$ at which STAS is applied. As shown in Figure \ref{fig:m}, steering at an intermediate layer ($m{=}9$) achieves the best performance on both metrics. 
In contrast, applying STAS at a shallower layer ($m{=}4$) or a deeper layer ($m{=}14$) leads to weaker results.
This suggests that intermediate layers provide a more suitable representation for our method, while steering too early or too late in the network is less effective.

\begin{figure*}[t]
\centering
\includegraphics[width=0.9\textwidth]{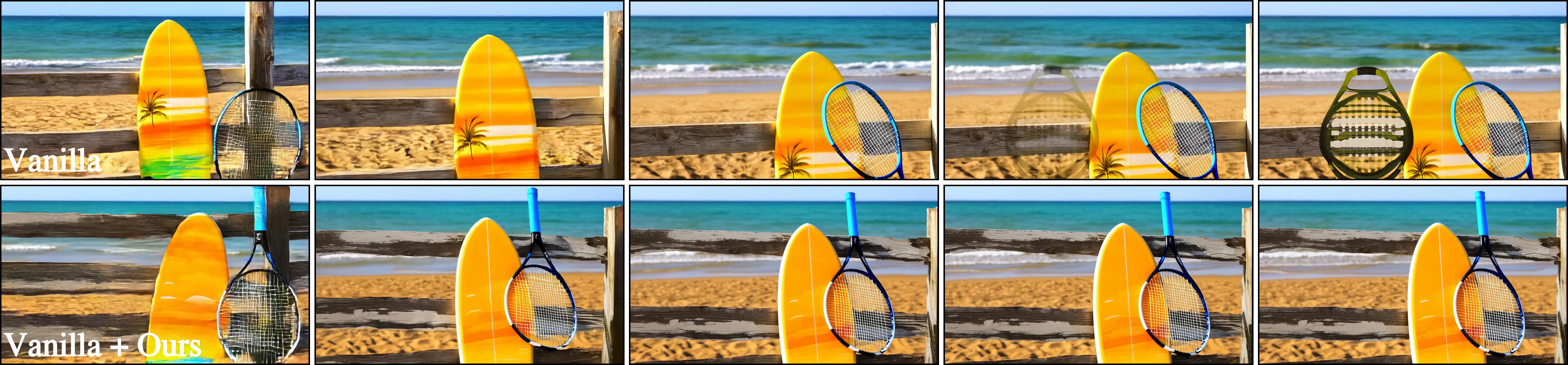}
\caption{\textbf{A failure case of Dynamic Degree evaluation.}
Matched frames for ``a surfboard and a tennis racket''. In the
baseline (top), the tennis racket disappears in the early frames and
later reappears together with an additional green racket.
Dynamic Degree counts these identity changes as motion and predicts
\texttt{True}. With STAS (bottom), one racket persists across all
frames and the prediction changes to \texttt{False}.}
\label{fig:dynamic_degree_limitation}
\end{figure*}

\subsection{Additional Qualitative Results}

\begin{figure}[ht]
    \centering
    \includegraphics[width=0.995\textwidth]{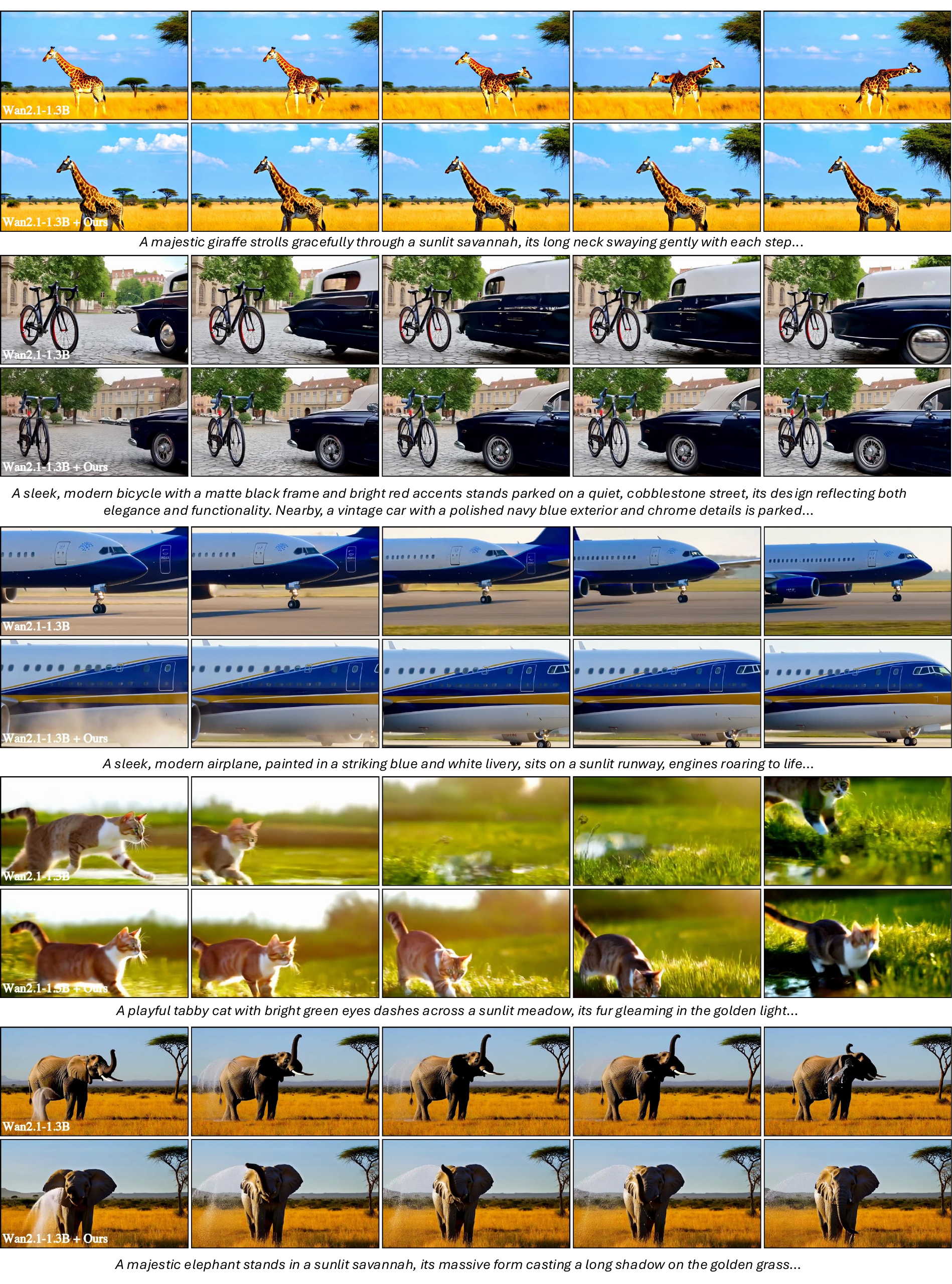}
    \caption{\textbf{More qualitative comparisons with and without STAS on Wan2.1-1.3B.}}
    \label{fig:app_wan2.1_1}
\end{figure}

\begin{figure}[ht]
    \centering
    \includegraphics[width=0.995\textwidth]{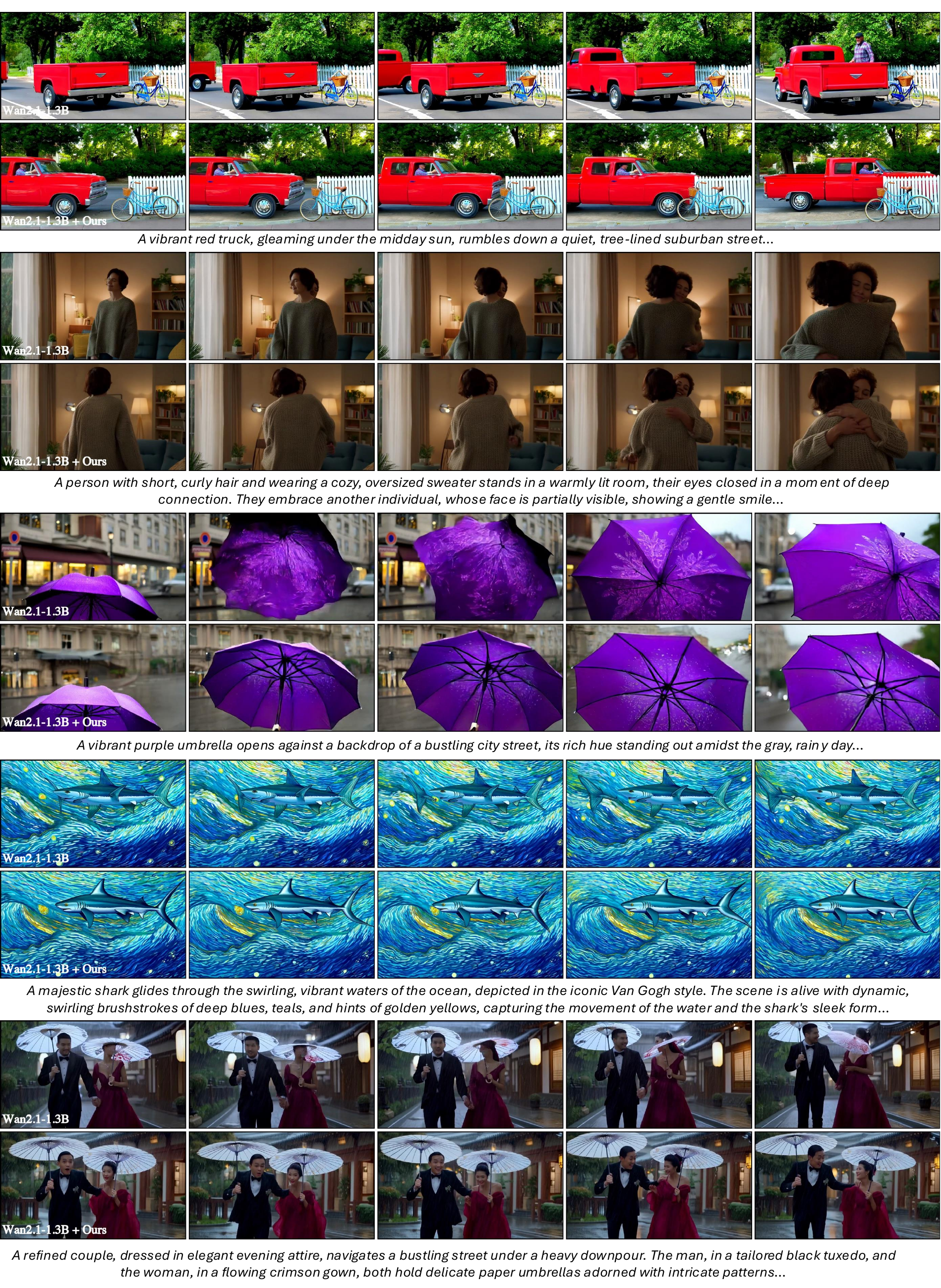}
    \caption{\textbf{More qualitative comparisons with and without STAS on Wan2.1-1.3B.}}
    \label{fig:app_wan2.1_2}
\end{figure}

\begin{figure}[ht]
    \centering
    \includegraphics[width=0.995\textwidth]{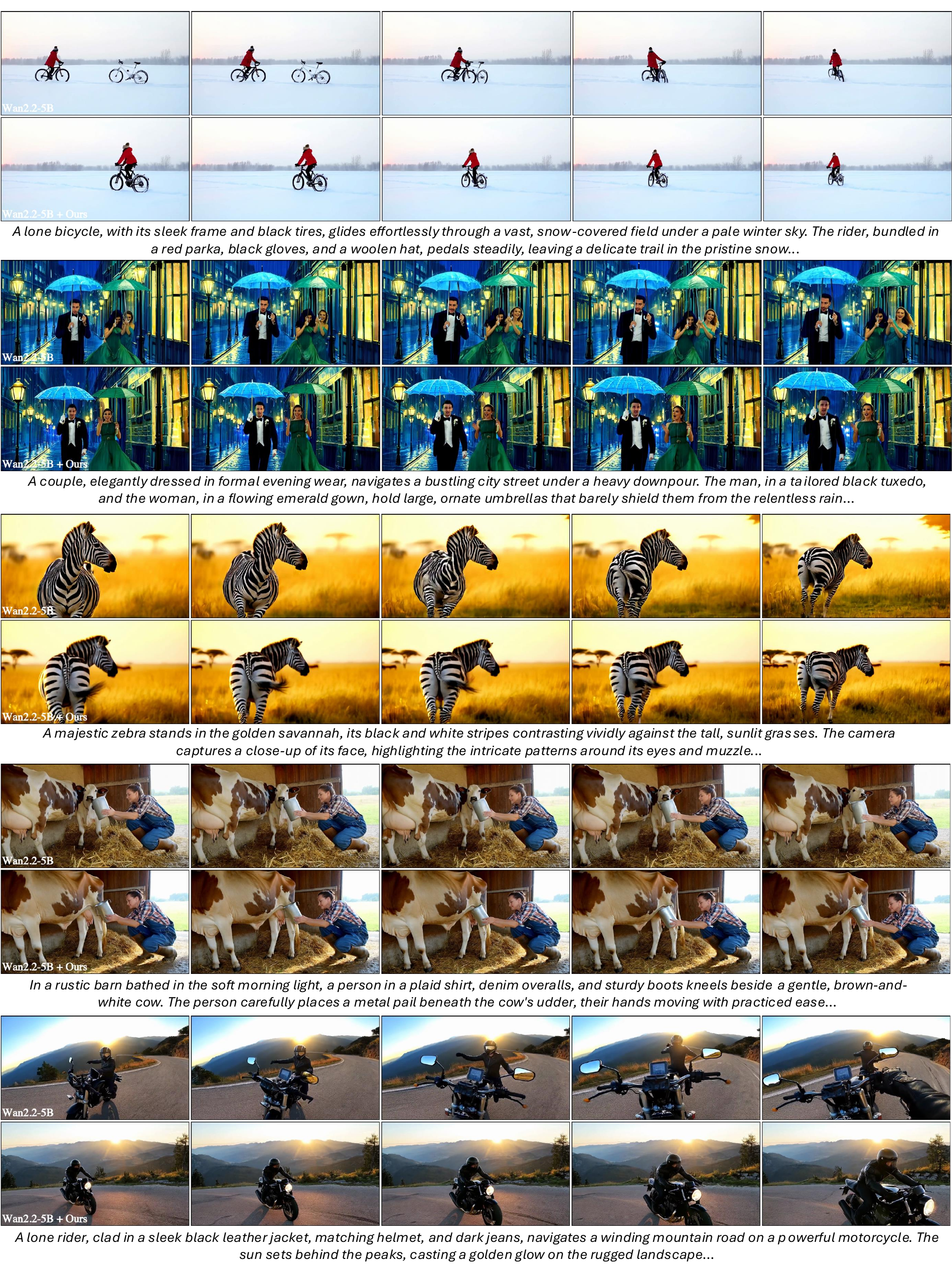}
    \caption{\textbf{More qualitative comparisons with and without STAS on Wan2.2-5B.}}
    \label{fig:app_wan2.2_1}
\end{figure}

\begin{figure}[ht]
    \centering
    \includegraphics[width=0.995\textwidth]{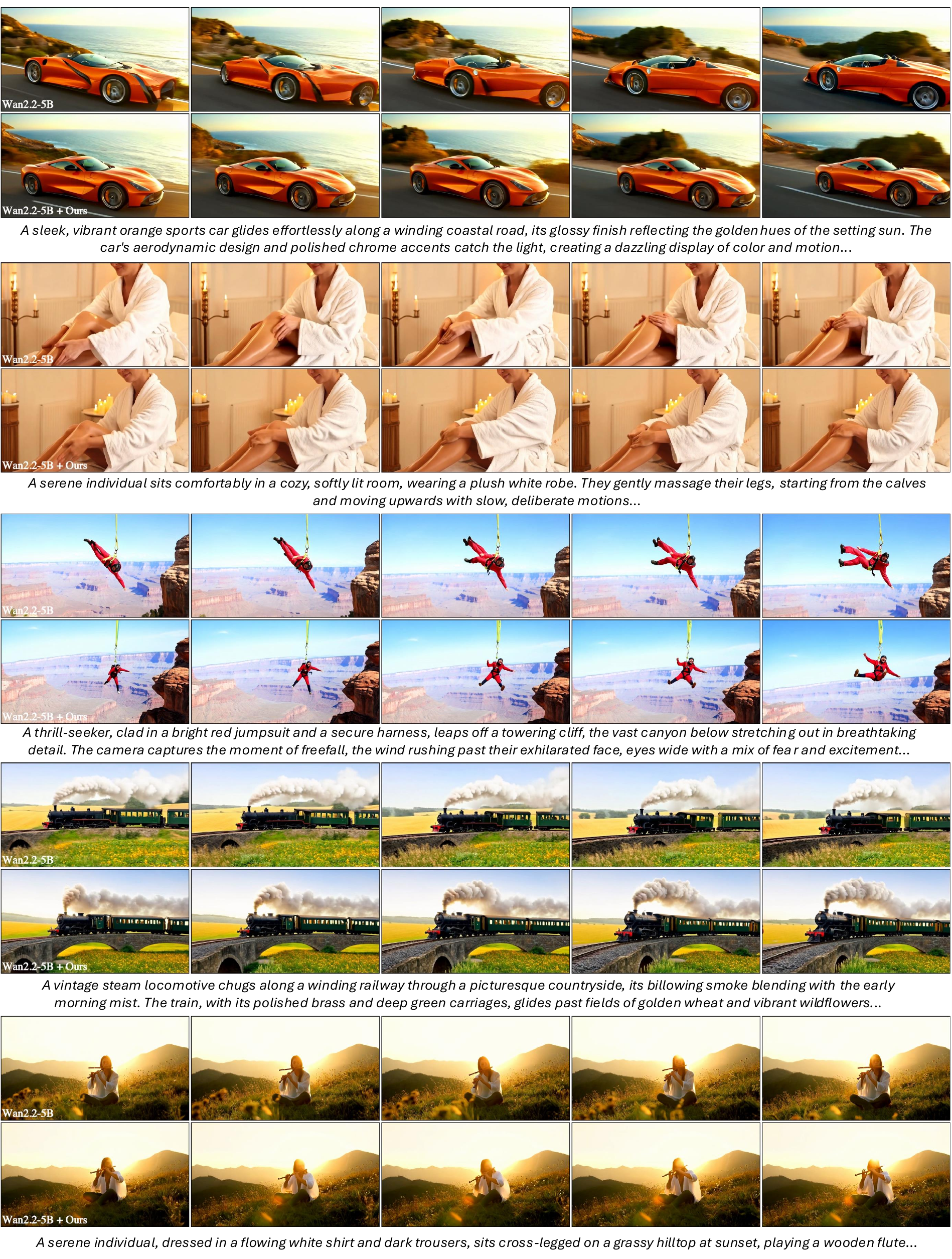}
    \caption{\textbf{More qualitative comparisons with and without STAS on Wan2.2-5B.}}
    \label{fig:app_wan2.2_2}
\end{figure}

\begin{figure}[ht]
    \centering
    \includegraphics[width=0.995\textwidth]{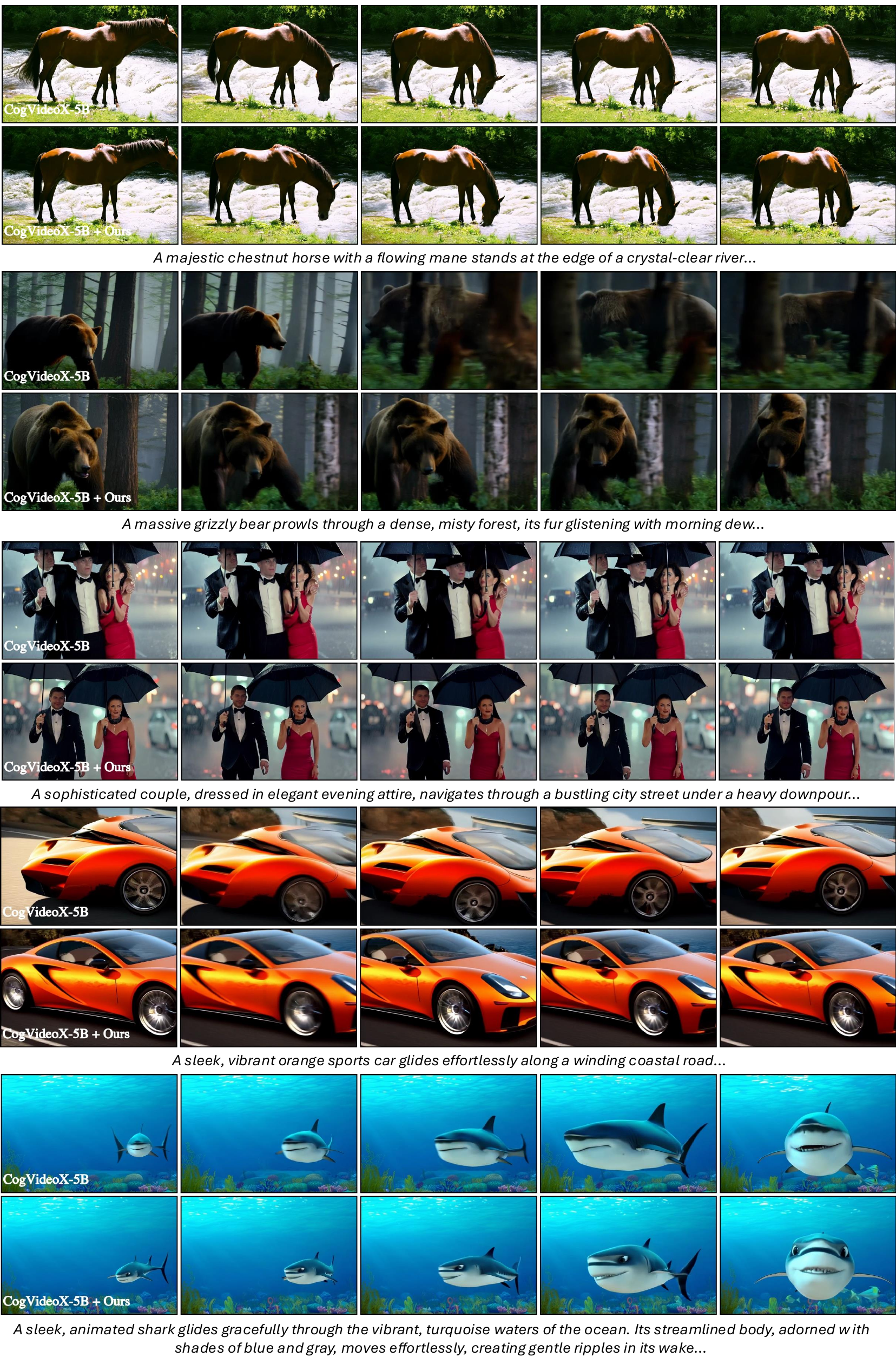}
    \caption{\textbf{More qualitative comparisons with and without STAS on CogvideoX-5B.}}
    \label{fig:app_cog1}
\end{figure}

We provide additional qualitative comparisons across different base models in Figures \ref{fig:app_wan2.1_1}, \ref{fig:app_wan2.1_2}, \ref{fig:app_wan2.2_1}, \ref{fig:app_wan2.2_2}, \ref{fig:app_cog1}, \ref{fig:app_cog2}, \ref{fig:app_ltx1} and \ref{fig:app_ltx2}. The example videos are also provided in the supplementary materials.

\section{Additional Discussion of Limitations}
\label{app:limitations}

Threshold-based metric like Dynamic Degree in Vbench can misinterpret temporal artifacts as
genuine motion. Figure~\ref{fig:dynamic_degree_limitation} shows matched
frames from the baseline and STAS for the prompt ``a surfboard and a
tennis racket''. The baseline exhibits both failure modes that inflate
frame-to-frame variation: the tennis racket first disappears over the
early frames, then returns later in the sequence together with an
additional green racket. Neither change represents meaningful
motion and both arise from inconsistent object identity. Dynamic Degree
nevertheless counts this variation as dynamics and predicts
\texttt{True}. With STAS, a single racket persists across the entire
sequence, the artifact-driven variation disappears, and the sample
falls below the metric's decision threshold, changing the prediction
to \texttt{False}. This transition therefore reflects the removal of artifact-driven variation rather
than of genuine motion.

\section{AI Use Statement}
In this work, we used generative AI tools for writing aid for grammar, wording, and formatting suggestions.
We have not used generative AI tools for autonomous generation of experimental evidence or figure drawing, and tasks involving AI-generated or simulated study participants are not applicable to this work. Additionally, we used generative AI tools for drafting and debugging analysis and visualization code. We have reviewed all AI-assisted work. AI-assisted code was manually inspected, executed, and verified against the underlying data by the authors. We take responsibility for the final content of this work,
including text, claims or artifacts produced with the aid of generative AI.

\clearpage

\begin{figure}[ht]
    \centering
    \includegraphics[width=0.995\textwidth]{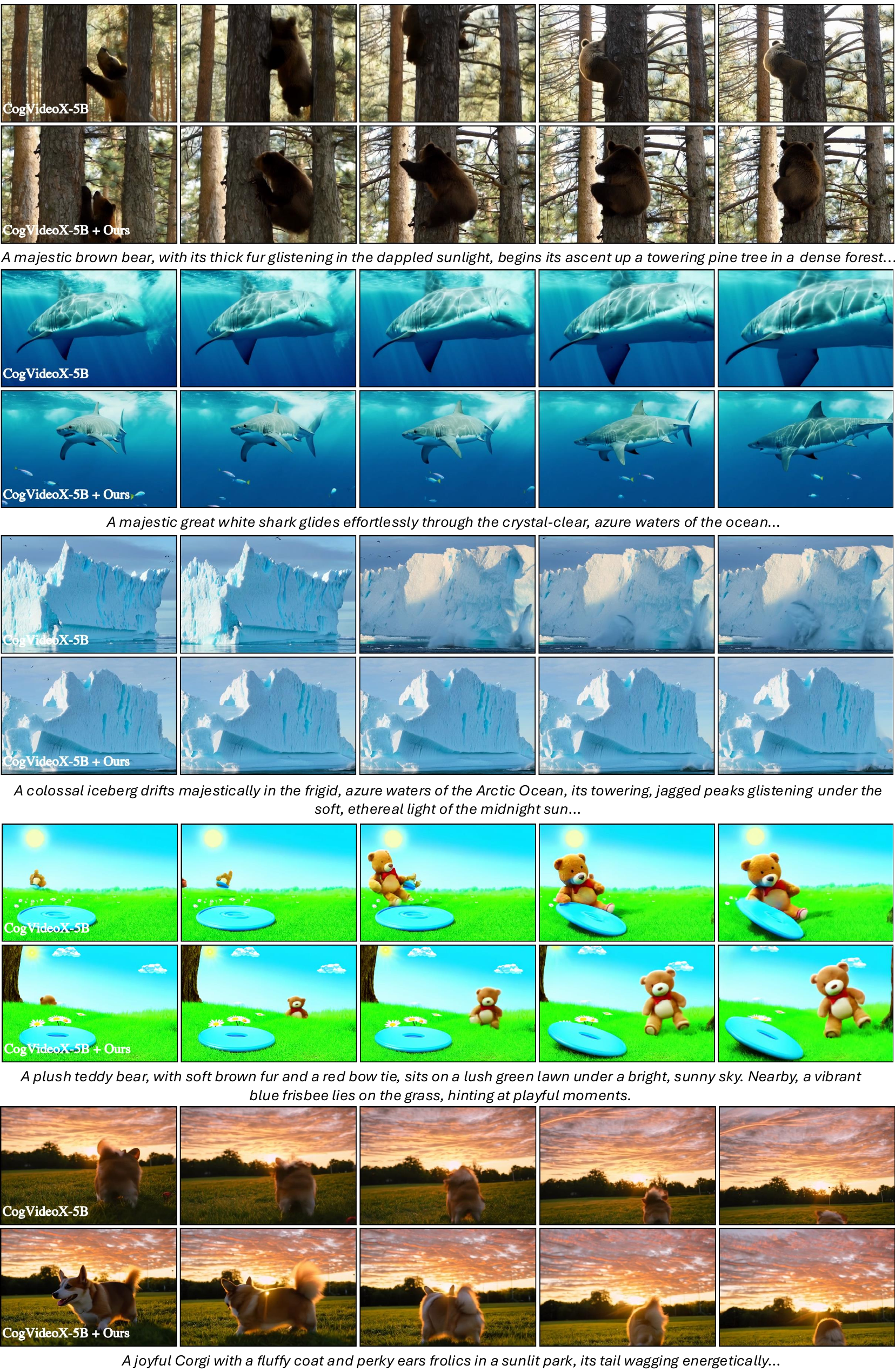}
    \caption{\textbf{More qualitative comparisons with and without STAS on CogvideoX-5B.}}
    \label{fig:app_cog2}
\end{figure}

\begin{figure}[ht]
    \centering
    \includegraphics[width=0.995\textwidth]{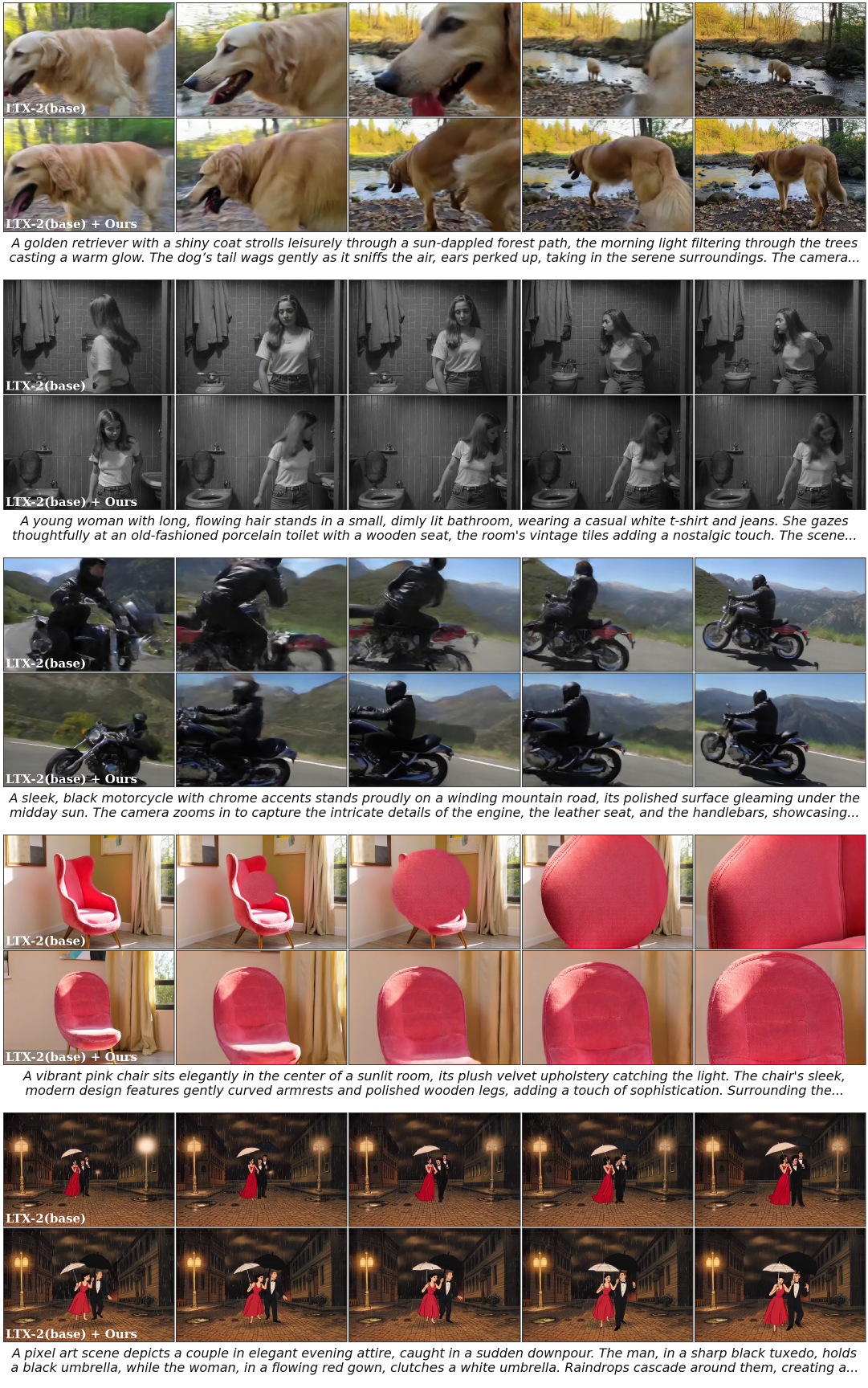}
    \caption{\textbf{More qualitative comparisons with and without STAS on LTX2 (base).}}
    \label{fig:app_ltx1}
\end{figure}

\begin{figure}[ht]
    \centering
    \includegraphics[width=0.995\textwidth]{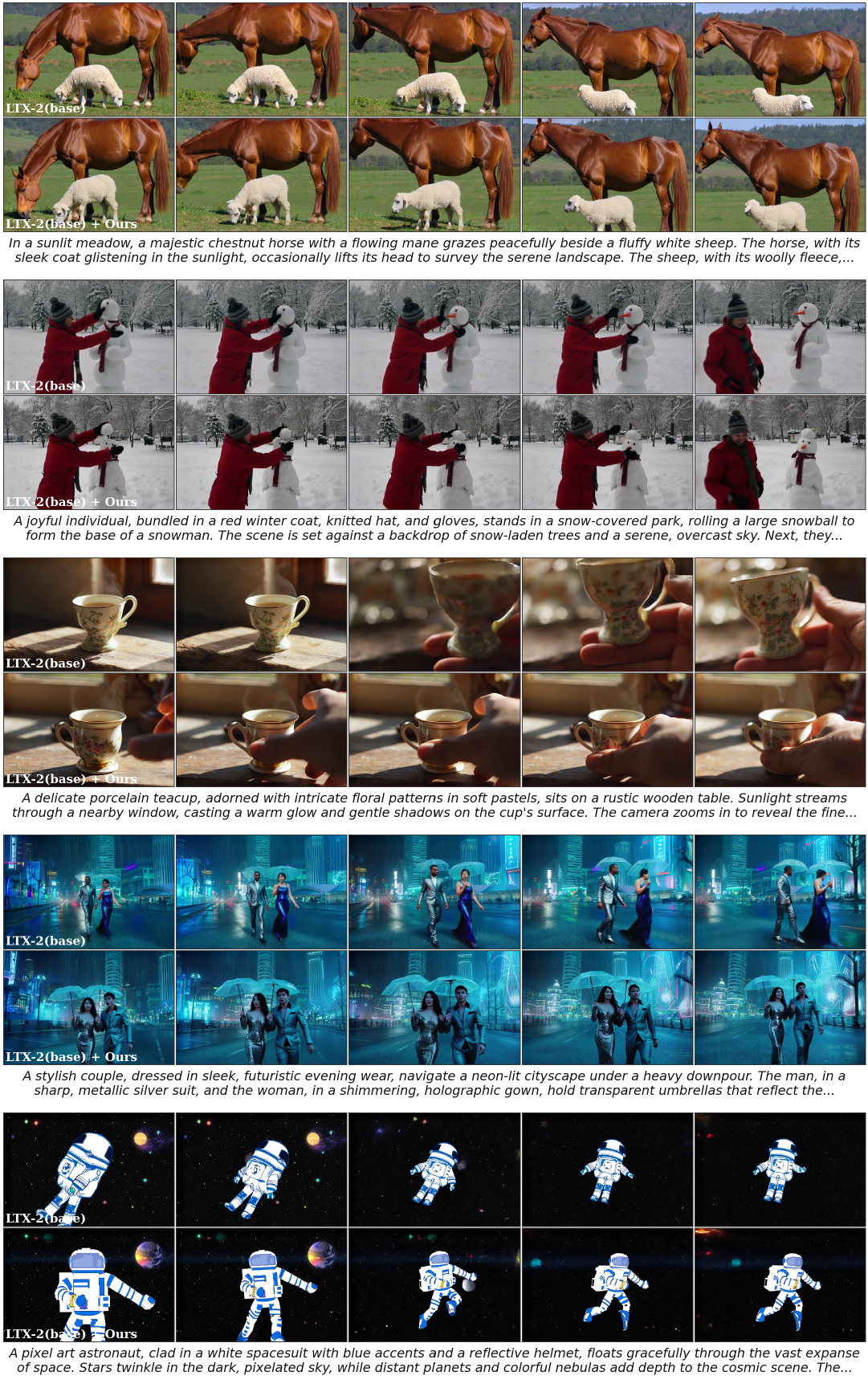}
    \caption{\textbf{More qualitative comparisons with and without STAS on LTX2 (base).}}
    \label{fig:app_ltx2}
\end{figure}

\end{document}